\PassOptionsToPackage{table}{xcolor}
\documentclass[lettersize,journal]{IEEEtran}
\usepackage{amsmath,amsfonts}
\usepackage{array}
\usepackage[caption=false,font=normalsize,labelfont=sf,textfont=sf]{subfig}
\usepackage{textcomp}
\usepackage{stfloats}
\usepackage{url}
\usepackage[capitalise]{cleveref}
\usepackage{verbatim}
\usepackage{graphicx}
\usepackage{tikz}
\usepackage{tikz-network}
\usepackage{cite}
\usepackage{diagbox}
\usepackage[ruled, lined, linesnumbered, commentsnumbered, longend]{algorithm2e}
\begin{document}

\title{Less Is More - \\ On the Importance of Sparsification for Transformers and Graph Neural Networks for TSP }

\author{Attila Lischka, Jiaming Wu, Rafael Basso, Morteza Haghir Chehreghani, Balázs Kulcsár

\thanks{Attila Lischka, Jiaming Wu, Morteza Haghir Chehreghani, and Balázs Kulcsár are with Chalmers University of Technology.}
\thanks{Rafael Basso is with Volvo Group Trucks Technology.}
\thanks{contact: \{lischka, jiaming.wu, morteza.chehreghani, kulcsar\}@chalmers.se, rafael.basso@volvo.com}
}

\maketitle

\begin{abstract}
Most of the recent studies tackling routing problems like the Traveling Salesman Problem (TSP) with machine learning use a transformer or Graph Neural Network (GNN) based encoder architecture. 
However, many of them apply these encoders naively by allowing them to aggregate information over the whole TSP instances.
We, on the other hand, propose a data preprocessing method that allows the encoders to focus on the most relevant parts of the TSP instances only. 
In particular, we propose \textit{graph sparsification} for TSP graph representations passed to GNNs and \textit{attention masking} for TSP instances passed to transformers where the masks correspond to the adjacency matrices of the sparse TSP graph representations.  
Furthermore, we propose ensembles of different sparsification levels allowing models to focus on the most promising parts while also allowing information flow between all nodes of a TSP instance. 
In the experimental studies, we show that for GNNs appropriate sparsification and ensembles of different sparsification levels lead to substantial performance increases of the overall architecture.
We also design a new, state-of-the-art transformer encoder with ensembles of attention masking. These transformers increase model performance from a gap of $0.16\%$ to $0.10\%$ for TSP instances of size 100 and from $0.02\%$ to $0.00\%$ for TSP instances of size 50.
\end{abstract}


\section{Introduction}
\IEEEPARstart{G}{raph} neural networks (GNNs; see \cref{gnn}) have emerged as a powerful architecture when dealing with graph structured data like molecules, social networks or traffic models \cite{wu2020comprehensive} in recent years.
Moreover, transformer models have achieved state-of-the-art performance in sequence-to-sequence tasks like machine translation \cite{vaswani2017attention}.
Concurrently, many studies deal with routing problems like the traveling salesman problem (TSP; see \cref{tsp}) or the capacitated vehicle routing problem (CVRP) using machine learning (see \cref{related_work}).
Despite the fact that many of these studies take completely different approaches and learning paradigms, often their architectures require some sort of \textit{encoder} in their framework which produces vector representations for the nodes in the problem instance.
GNNs and transformers (which are closely related to GNNs, see \cite{joshi2020transformers}) have been successfully used as encoders for routing problems in various settings since routing problems can easily be interpreted as graph problems.
However, so far, most of the studies using these architectures as encoders naively perform their models on the whole \textit{dense} TSP graph. %
These dense graph representations reflect that, e.g. in TSP, it is possible to travel from any city to any other city in the TSP instance.
However, GNNs are known to take advantage of the underlying graph structure of an input graph to produce node encodings \cite{xu2018powerful, morris2019weisfeiler}. Dense graph representations have no exploitable structure as they allow an information flow between any pair of nodes in the message passing operations of a GNN. This means node embeddings will contain information of far-away, irrelevant nodes. Even worse, as all nodes are connected to all other nodes, they will share the exact same information in each message passing iteration, resulting in similar embeddings for all nodes. Therefore, afterwards, it will be difficult for the “decoder” part of the entire architecture to discriminate meaningfully between the different node embeddings.

\begin{figure*}[!ht]
\centering
\subfloat[GNN on optimal TSP edges only]{
  \resizebox{7cm}{!}{%
  \begin{tikzpicture}
    \node[shape=circle,draw=white] (9) at (3,-0.25) {};
    \node[shape=circle,draw=white] (10) at (4.5,-0.25) {};
    \Edge[Direct](9)(10)
    \node[shape=circle,draw=black, label=left:{$ \tiny \begin{pmatrix} \color{red}{1} \\ 0 \\ 0 \\ 0 \\ 0\end{pmatrix}$}] (1) at (0,0.5) {1};
    \node[shape=circle,draw=black, label=above:{$\setlength\arraycolsep{1pt} \tiny \begin{pmatrix}0 & \color{red}{1} & 0 & 0 & 0 \end{pmatrix}^\top$}] (2) at (1.5,1) {2};
    \node[shape=circle,draw=black, label=left:{$\tiny \begin{pmatrix}0 \\ 0 \\ \color{red}{1} \\ 0 \\ 0\end{pmatrix}$}] (3) at (0,-1) {3};
    \node[shape=circle,draw=black, label=right:{$\tiny \begin{pmatrix}0 \\ 0 \\ 0 \\ \color{red}{1} \\ 0 \end{pmatrix}$}] (4) at (2,-0.25) {4};
    \node[shape=circle,draw=black, label=below:{$\setlength\arraycolsep{1pt} \tiny \begin{pmatrix}0 & 0 & 0 & 0 & \color{red}{1}\end{pmatrix}^\top$}] (5) at (1.5,-1.5) {5};
    \draw[latex'-latex', thick] (1) -- (2);
    \draw[latex'-latex', thick] (1) -- (3);
    \draw[latex'-latex', thick] (4) -- (2);
    \draw[latex'-latex', thick] (4) -- (5);
    \draw[latex'-latex', thick] (3) -- (5);
    \node[shape=circle,draw=black, label=left:{$\tiny \begin{pmatrix} \color{blue}{2} \\ \color{red}{1} \\ \color{red}{1} \\ 0 \\ 0\end{pmatrix}$}] (6) at (5.5,0.5) {1};
    \node[shape=circle,draw=black, label=above:{$\setlength\arraycolsep{1pt}  \tiny \begin{pmatrix}\color{red}{1} & \color{blue}{2} & 0 & \color{red}{1} & 0 \end{pmatrix}^\top$}] (7) at (7,1) {2};
    \node[shape=circle,draw=black, label=left:{$\tiny \begin{pmatrix}\color{red}{1} \\ 0 \\ \color{blue}{2} \\ 0 \\ \color{red}{1}\end{pmatrix}$}] (8) at (5.5,-1) {3};
    \node[shape=circle,draw=black, label=right:{$\tiny \begin{pmatrix}0 \\ \color{red}{1} \\ 0 \\ \color{blue}{2} \\ \color{red}{1} \end{pmatrix}$}] (9) at (7.5,-0.25) {4};
    \node[shape=circle,draw=black, label=below:{$\setlength\arraycolsep{1pt} \tiny \begin{pmatrix}0 & 0 & \color{red}{1} & \color{red}{1} & \color{blue}{2} \end{pmatrix}^\top$}] (10) at (7,-1.5) {5};
    \Edge(6)(7)
    \Edge(6)(8)
    \Edge(7)(9)
    \Edge(9)(10)
    \Edge(8)(10)
  \end{tikzpicture}
  }
  \label{fig:sparse_gnn}
  }
  \hfil
  \subfloat[GNN on complete graph]{
  \resizebox{7cm}{!}{%
  \begin{tikzpicture}
    \node[shape=circle,draw=white] (9) at (3.0,-0.25) {};
    \node[shape=circle,draw=white] (10) at (4.5,-0.25) {};
    \Edge[Direct](9)(10)
    \node[shape=circle,draw=black, label=left:{$\tiny \begin{pmatrix}\color{red}{1} \\ 0 \\ 0 \\ 0 \\ 0\end{pmatrix}$}] (1) at (0,0.5) {1};
    \node[shape=circle,draw=black, label=above:{$ \setlength\arraycolsep{1pt} \tiny \begin{pmatrix}0 & \color{red}{1} & 0 & 0 & 0 \end{pmatrix}^\top$}] (2) at (1.5,1) {2};
    \node[shape=circle,draw=black, label=left:{$\tiny \begin{pmatrix}0 \\ 0 \\ \color{red}{1} \\ 0 \\ 0\end{pmatrix}$}] (3) at (0,-1) {3};
    \node[shape=circle,draw=black, label=right:{$\tiny \begin{pmatrix}0 \\ 0 \\ 0 \\ \color{red}{1} \\ 0 \end{pmatrix}$}] (4) at (2,-0.25) {4};
    \node[shape=circle,draw=black, label=below:{$ \setlength\arraycolsep{1pt} \tiny \begin{pmatrix}0 & 0 & 0 & 0 & \color{red}{1}\end{pmatrix}^\top$}] (5) at (1.5,-1.5) {5};
    \draw[latex'-latex', thick] (1) -- (2);
    \draw[latex'-latex', thick] (1) -- (3);
    \draw[latex'-latex', thick] (4) -- (2);
    \draw[latex'-latex', thick] (4) -- (5);
    \draw[latex'-latex', thick] (3) -- (5);
    \draw[latex'-latex', thick] (1) -- (4);
    \draw[latex'-latex', thick] (1) -- (5);
    \draw[latex'-latex', thick] (3) -- (2);
    \draw[latex'-latex', thick] (5) -- (2);
    \draw[latex'-latex', thick] (3) -- (4);
    \node[shape=circle,draw=black, label=left:{$\tiny \begin{pmatrix} \color{blue}{2} \\ \color{red}{1} \\ \color{red}{1} \\ \color{red}{1} \\ \color{red}{1}\end{pmatrix}$}] (6) at (5.5,0.5) {1};
    \node[shape=circle,draw=black, label=above:{$\setlength\arraycolsep{1pt} \tiny \begin{pmatrix}\color{red}{1} & \color{blue}{2} & \color{red}{1} & \color{red}{1} & \color{red}{1} \end{pmatrix}^\top$}] (7) at (7,1) {2};
    \node[shape=circle,draw=black, label=left:{$\tiny \begin{pmatrix}\color{red}{1} \\ \color{red}{1} \\ \color{blue}{2} \\ \color{red}{1} \\ \color{red}{1}\end{pmatrix}$}] (8) at (5.5,-1) {3};
    \node[shape=circle,draw=black, label=right:{$\tiny \begin{pmatrix} \color{red}{1} \\ \color{red}{1} \\ \color{red}{1} \\ \color{blue}{2} \\ \color{red}{1} \end{pmatrix}$}] (9) at (7.5,-0.25) {4};
    \node[shape=circle,draw=black, label=below:{$\setlength\arraycolsep{1pt} \tiny \begin{pmatrix}\color{red}{1} & \color{red}{1} & \color{red}{1} & \color{red}{1} &  \color{blue}{2}\end{pmatrix}^\top$}] (10) at (7,-1.5) {5};
    \Edge(6)(7)
    \Edge(6)(8)
    \Edge(7)(9)
    \Edge(9)(10)
    \Edge(8)(10)
    \Edge(6)(9)
    \Edge(6)(10)
    \Edge(7)(8)
    \Edge(7)(10)
    \Edge(8)(9)
  \end{tikzpicture}
  }
  \label{fig:dense_gnn}
    }
\caption{Message passing on GNNs - importance of sparsification}
\label{fig:gnn_sparsification}
\end{figure*}
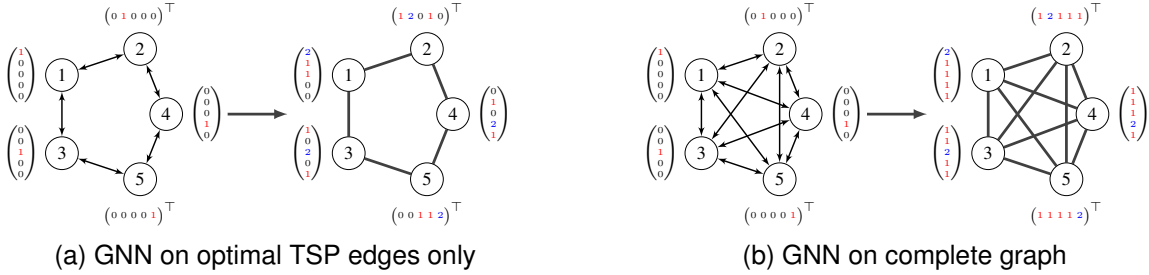

We visualize this issue in \cref{fig:gnn_sparsification}.
In \cref{fig:sparse_gnn}, the only edges in the graph are the ones that correspond to the actual TSP solution, whereas \cref{fig:dense_gnn} represents a complete graph.
In the figure, we consider a simple GNN that updates the feature vectors by adding the feature vectors of all nodes in the neighborhood (including itself) to the previous feature vectors.
Furthermore, each node has a unique initial encoding (e.g., a one-hot encoding).
After initializing the node embeddings with this encoding in a first step (left halves of \cref{fig:sparse_gnn} and \cref{fig:dense_gnn}), message passing is performed as the next step in the GNN. 
We show this in the right halves of \cref{fig:sparse_gnn} and \cref{fig:dense_gnn}, where the encodings after one round of message passing are shown.
Note that each node embedding now has a 2-entry where it had a 1 entry before and additional 1-entries reflecting its neighbors.
We can see that in \cref{fig:sparse_gnn}, after only one message passing update, it would be possible to decode the whole solution given the updated feature vectors.
After choosing a feature vector to start from, the next feature vector to visit should be one that has not been visited yet (easily achievable by masking) and whose entry in the current feature vector corresponds to a $1$. 
This next node can be found by simply iterating over the other nodes' feature vectors and choosing a node whose feature vector has entry $2$ at the position where the current node's feature vector has entry $1$. 
For \cref{fig:dense_gnn}, this would not be possible, on the other hand, as in this setting the updated feature vectors do not contain any exploitable information based on which the decoder could choose the next node to select since all entries that previously were $0$ are now $1$.

To overcome the described limitations of dense TSP graph representations for GNN (and transformer) encoders, in this work, we investigate the influence of preprocessing TSP graphs. This preprocessing means deleting edges, which are unpromising to be part of the optimal TSP solution, before passing the graphs to a GNN encoder. 
Similarly, we can mask out the attention scores of TSP instances passed to a transformer encoder based on the adjacency matrix of the sparsified TSP graph.
This preprocessing allows the encoder to focus on the relevant parts of the problem and, therefore, to produce better embeddings for further processing in the overall framework. %
Moreover, to mitigate the possibility of deleting optimal edges in a too far-reaching sparsification, we propose ensembles of different sparsification levels which allows an encoder to simultaneously focus on the most promising edges in a TSP instance while still allowing an information flow between all the components.
In particular, we provide the following contributions.
\begin{itemize}
    \item We propose two data preprocessing methods for TSP that, by deleting unpromising edges, make the corresponding TSP instances sparse, allowing GNNs and transfomer encoders to focus on the most relevant parts of the problem instances.
    Firstly, we propose the simple $k$-nearest neighbors heuristic and, secondly, 1-Trees, a minimum spanning tree based approach which was previously used successfully in the initialization of the powerful LKH algorithm \cite{Helsgaun2000AnEI}. We then compare both methods w.r.t. their capability for keeping the optimal edges in the sparse representation of a TSP instance.
    \item To demonstrate the significance of sparsification in practice, we evaluate our sparsification methods with two different GNN architectures, namely, \textit{Graph Attention Networks} (GAT) and \textit{Graph Convolutional Networks} (GCN) on two different data distributions and show that performance improves in all settings when sparsification strategies are applied. We show, for example, that the optimality gap achieved when training a GAT encoder on 20000 TSP instances of size 100 on the uniform data distribution decreases from $15.66\%$ to $0.7\%$ when sparsifying the instances properly - a $22 \times$ improvement. On the same dataset, the performance of the GCN improves by a factor of over $2 \times$ from a gap of $2.35\%$ to $0.94\%$. We further propose ensembles of different sparsification levels which leads to a gap of $0.7\%$ for GATs (same performance as the best individual sparsification level) and $0.77\%$ for GCNs (better than any single sparsification level). An overview of how the sparsification is incorporated in the overall learning framework is given in \cref{fig:flowchart}.
    \item We implement an attention masking mechanism for transformers that reflects the 1-Tree based edge sparsification for GNNs, indicating that the sparsification process is relevant for transformer architectures as well. We evaluate this masking for different sparsification levels as well as ensembles consisting of several sparsification levels and show that the performance of the ensemble improves (decreases) the optimality gap from $0.16\%$ (no attention masking) to $0.10\%$ leading to the new state-of-the-art encoder for learning based TSP solvers of the ``encoder-decoder'' category (compare \cref{related_work}) for TSP instances of size 100. Further, this ensemble transformer achieves a gap of $0.00\%$ for TSP instances of size 50 down from $0.02\%$ from the previously best ``encoder-decoder'' framework.
\end{itemize}

To the best of our knowledge, our work is the first to investigate the importance of graph sparsification as a form of data preprocessing for the TSP. 
We believe that our preprocessing can, together with suitable encoders, be adapted for many existing machine learning frameworks dealing with the TSP, or even beyond, i.e., to many other combinatorial optimization problems where initial conditions are of paramount importance.

The remainder of the paper is organized as follows: 
first, we present the related work in \cref{related_work}. Then, we introduce the relevant technical preliminaries in \cref{preliminaries}. Next, we describe our methodology in \cref{methodology}, before presenting our experimental studies in \cref{total_experiments}. Finally, we conclude in \cref{conclusion}.

\section{Related Work} \label{related_work}

\subsection{Learn to Route}
Prior work addressing routing problems such as the TSP or CVRP with machine learning approaches can be grouped by different characteristics. 
One possibility is, for example, to group them by their learning paradigm (i.e., supervised learning, reinforcement learning, or even unsupervised learning).
Another possibility is to group them by the underlying neural architecture, typically GNNs, RNNs, or attention based models like transformers (or combinations of these architecture types).
In this paper, we stick to the most common categorization: Grouping them by the way machine learning is used within the overall framework.

\textbf{Encoder-decoder approaches}: Here, an encoder architecture produces embeddings for each of the nodes in the instance, which are then autoregressively selected by the decoder architecture to form a tour.
By masking invalid decisions, a feasible solution is ensured.
Examples are \cite{deudon2018learning, nazari2018reinforcement, kool2018attention, ma2019combinatorial, kwon2020pomo, xin2020multidecoder, jin2023pointerformer}.

\textbf{Search-based approaches}:
Here, the architecture is trained to learn a sort of cost metric which can later be transformed into a valid solution by a search algorithm.
For example, \cite{joshi2019efficient, fu2021, kool2022deep, min2023unsupervised}, learn \textit{edge probability} heatmaps which indicate how likely an edge is part of an optimal solution. \cite{qiu2022dimes} learn a similar edge score heatmap (which does not directly correspond to probabilities) which can later be used to guide a search algorithm.
\cite{hudson2021graph} learn to predict a regret score for each edge which is afterwards used in a guided local search.

\textbf{Improvement-based approaches}:
Here, the aim is to improve a given valid tour.
This is done either by using typical operation research improvement operators like $k$-opt (where $k$ edges are deleted and $k$ other edges added) like in \cite{d2020learning, lu2020learning, wu2021learning}, or by learning to select and/or optimize subproblems (i.e., for example, subtours or subpaths of the current solution) like in \cite{chen2019learning, kim2021learning, li2021learning, zong2022rbg, cheng2023select}.

We note that, typically, machine learning models that aim to solve routing problems like the TSP contain some sort of internal encoder architecture.
This is most obvious for encoder-decoder architectures. 
However, search-based approaches also contain some sort of encoder architecture:
In this setting, the encoder is the part of the architecture responsible for creating hidden feature vectors for all the nodes in the instance which are later used to create the edge scores.
In improvement-based approaches, it is less obvious, but even these frameworks use (or could use) an internal encoder to produce meaningful representations (i.e., embeddings) for the nodes as it is done, e.g., in \hbox{\cite{d2020learning}}.
We emphasize that these encoders can and have been trained with different learning techniques: e.g., \cite{joshi2019efficient} uses supervised learning, \cite{kool2018attention} uses reinforcement learning and \hbox{\cite{min2023unsupervised}} was the first work to even train in an unsupervised learning manner.

In our work, we propose a method to make GNN and transformer encoders, that are used in the different learning settings mentioned above, more powerful when employed to compute embeddings for the TSP in learning tasks.
Therefore, our contribution is orthogonal to the above categorization as it is applicable in all settings.
In particular, we point out that in e.g. \cite{joshi2019efficient, d2020learning, fu2021, kool2022deep, min2023unsupervised} GNNs and in \cite{kool2018attention, kwon2020pomo, jin2023pointerformer} transformers were used as encoders operating on dense graph representations of the TSP. 
These studies are from the different categorizations presented above and we hypothesize that all of these architectures could achieve better performances when using our proposed methods to sparsify the graphs first.
We emphasize that \cite{fu2021} which deals with scalability, aims to solve large TSP instances and points out that our data preprocessing is applicable in this setting as well, e.g. by sparsifying the sampled subgraphs in the overall framework (see \cite{fu2021} for details).

\subsection{Sparsification for Routing}
We further note that, until now, most studies that use GNNs as encoders for learning routing problems completely ignore the limitations of dense TSP graphs (outlined in the Introduction).
Only \cite{xin2021neurolkh} and \cite{qiu2022dimes} try to make the underlying graph sparse by using the $k$-nearest neighbor ($k$-nn) heuristic, where each node is only connected to its $k$ closest neighbors (using Euclidean distance).
We note that the motivation of \cite{qiu2022dimes} to make graphs sparse is related to runtime (computational aspects) as the number of edges in a complete graph grows quadratically in the number of nodes $n$. 
By using $k$-nn to make the graphs sparse, the number of edges can be reduced from $\mathcal{O}(n^2)$ to $\mathcal{O}(kn)$. 
\cite{xin2021neurolkh} acknowledges the importance of sparsification for effective training but chooses a fixed $k=20$ and does not investigate the level of sparsification.
Moreover, \cite{xin2021neurolkh} does not propose sparsification as a general tool to create more powerful GNN encoders, but to predict edge scores and node penalties in their specific NeuroLKH framework.

\section{Preliminaries} \label{preliminaries}
\subsection{Graph Neural Networks} \label{gnn}

Graph Neural Networks (GNNs) are a class of neural architectures that allow for capturing the underlying graph structure of their input.
A GNN computes feature vectors for all the nodes in the input graph.
In each layer, a node $i$ receives information from all other nodes $j$ it is connected to via an edge (called \textit{neighborhood} $N(i)$). 
This process is called \textit{message passing}.
A node's feature vector is updated by aggregating the set of received messages.
The aggregated information is typically normalized, passed through a neural layer and added to the old node vector representation.
Additional activation functions applied to the intermediate outputs lead to non-linearities and the capability to learn complex functions.

In this work, we focus on two different GNN types: A version of the Graph Convolutional Network (GCN) (originally introduced by \cite{kipf2016semi}) which was adapted by \cite{morris2019weisfeiler}:
$$x_i^\ell = \mathbf{W_1}^\ell \cdot x_i^{\ell-1} + \mathbf{W_2}^\ell \cdot \sum_{j \in N(i)} e_{j,i} \cdot x_j^{\ell-1},$$
where $x_i^\ell$ is the feature vector of node $i$ in the $\ell$th layer, $e_{j,i}$ is a normalization score indicating how important the feature vector of neighboring node $j$ is to node $i$, and $\mathbf{W_1}^\ell, \mathbf{W_2}^\ell$ are learnable weight matrices. 

Moreover, we use the version of the Graph Attention Network (GAT; originally introduced by \cite{velickovic2017graph}) proposed by \cite{brody2021attentive}, where the main difference compared to the GCN is that this architecture does not have fixed scores $e_{j,i}$ which are passed to the network, but these scores are learned by the GNN itself using the attention mechanism \cite{vaswani2017attention}.
Moreover, this GNN also allows us to use edge features (information for each edge in the graph) which can be used for learning.
\subsection{Travelling Salesman Problem} \label{tsp}
The Travelling Salesman Problem (TSP) is a combinatorial graph problem.
The problem consists of a set of cities (formally nodes or vertices) $V = \{1, 2, \dots, n\}$ and a distance or cost function $c: V \times V \mapsto \mathbb{R}$, which can, but does not need to be, symmetric.
The goal is to find a tour that visits every node exactly once and ends in the same city where it started (i.e. a Hamiltonian cycle) while minimizing the total traveled distance.
Typically, it is possible to travel from any city to any other city.
Therefore, from a graph-theoretic perspective, the problem then corresponds to a complete graph (without self-loops) $G=(V,E)$ where $V$ is defined as above and $E = \{(v,u) | u,v \in V, u \neq v\}$. 
In our setting, we assume a symmetric cost function that corresponds to the Euclidean distance between the nodes, given each node has a pair of coordinates $(x,y)$.

\section{Methodology} \label{methodology}

\subsection{Making the TSP Sparse: The Sparsification Process}
\begin{figure*}[!t]
\centering
\subfloat[Spasification by $k$-nn]{
  \includegraphics[width=.24\linewidth]{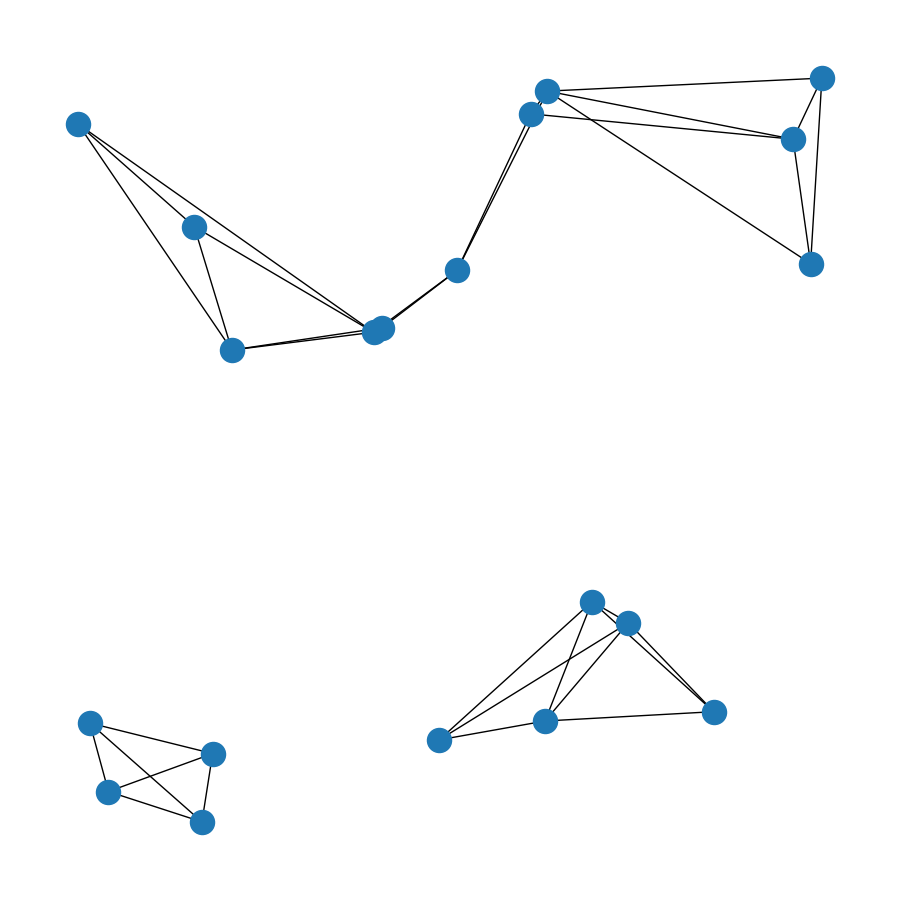}
  \label{fig:knn_sparse_small}
}
\hfil
\subfloat[Sparsification by 1-Tree]{
  \includegraphics[width=.24\linewidth]{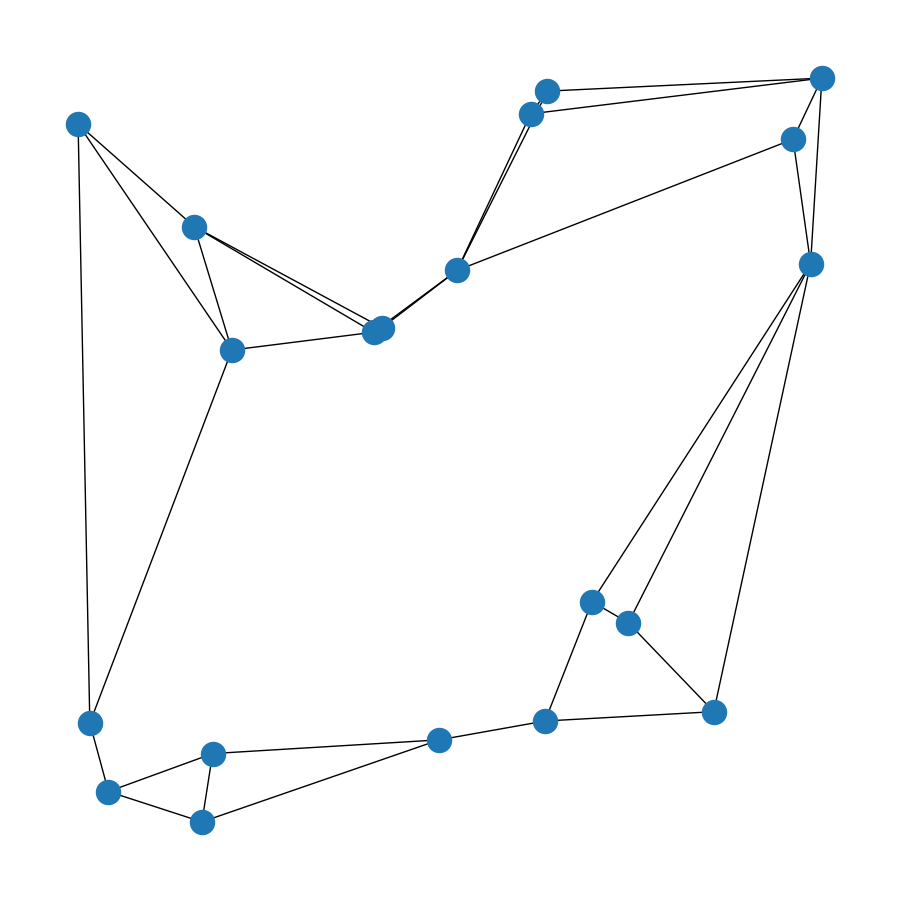}
  \label{fig:onetree_sparse_small}
}
\caption{Graph sparsified with $k$-nn and 1-Tree, keeping the 3 most promising edges for each node}
\label{fig:graphs_sparse}
\end{figure*}

As motivated in the Introduction, dense graphs can lead to information flooding when passed to GNNs, making it difficult to learn meaningful embeddings for the nodes.
Like it was hinted in \cite{xin2021neurolkh} and \cite{qiu2022dimes}, $k$-nn is a straightforward and easy heuristic to make graphs sparse.
However, this heuristic comes with a substantial theoretical drawback: the resulting sparse graph might not be connected.
This is obvious if the graph's nodes are clustered. 
If a graph contains two far-away clusters with $k+1$ nodes each, then $k$-nn will not result in a connected sparse graph. 
However, even with the data distributions assumed in most prior work on learning-based TSP, i.e. uniform and random distribution of the nodes in the unit square, unconnected graphs can occur if they are made sparse with $k$-nn (see \cref{fig:knn_sparse_small}).
This problem can of course be mitigated by increasing $k$, but this method yields losing structural information and the node embeddings get flooded with unnecessary information in the message passing steps again.
On the other hand, if the sparse graph is not connected, no information will flow between the different connected components in the message passing, making it difficult to learn embeddings that encode information of far away \textit{but relevant} nodes.

In a perfect setting, the edges in the sparse graph would correspond exactly to the TSP solution (recall \cref{fig:gnn_sparsification}).
By this, each node would exactly receive the information of its relevant neighbors.
Obviously, this is not practical in real cases.
Nevertheless, for a GNN to produce the best encodings possible on a sparse TSP graph, we expect it to:
\begin{enumerate}
    \item Have as few edges as possible while containing all (or at least as many as possible) edges from the optimal TSP solution.
    \item Be connected (i.e., the graph consists of one connected component only).
\end{enumerate}
Furthermore, we expect the sparsification step to be numerically fast.
To achieve all of these objectives, we propose to use the \textit{candidate set} of the LKH algorithm \cite{Helsgaun2000AnEI} as the edges in the sparse graphs.
The LKH algorithm performs $k$-opt moves on a given (suboptimal) solution to improve it until convergence. 
$k$-opt means that $k$ edges in the current solution are substitutes by $k$ edges not present in the current solution.
In order to restrict this search, the newly added edges must be included in the aforementioned \textit{candidate set}.
The LKH algorithm uses \textit{1-Trees}, a variant of minimum spanning trees (MSTs) which are modified by a subgradient optimization procedure to be ``closer'' to a TSP solution, to compute the candidate set (see \cite{Helsgaun2000AnEI} for details on how to compute the 1-Tree of a graph).
Note that a 1-Tree is assigned a cost: the sum of all edge costs in the 1-Tree. 
Therefore, there exists a cheapest 1-Tree (similar to a minimum spanning tree).
Furthermore, we can force an edge to be part of a 1-Tree and compute the remainder of the 1-Tree based on this first, given edge. 
This is again similar to constructing an MST with the additional constraint that a certain edge has to be part of the solution.
Note that 1-Trees with enforced edges have a cost that is at least as high as the cost of the cheapest 1-Tree.

Given a TSP instance $V$, the candidate set is generated as outlined in Algorithm \ref{alg:candset}.
The algorithm first computes $\alpha$ scores for all edges, indicating how expensive it is to enforce an edge in a 1-Tree compared to the cheapest 1-Tree possible.
Afterwards, ranking by the alpha scores, the best $k'$ edges (where $k'$ is a hyperparameter) for each node are kept in the candidate set.
\IncMargin{1em}
\begin{algorithm}[!t]
    \SetKwFunction{computeOneTree}{computeOneTree}
    \SetKwFunction{computeOneTreeForcingEdge}{computeOneTreeForcingEdge}
    \SetKwFunction{dict}{dict}
    \SetKwFunction{set}{set}
    \SetKwFunction{cost}{cost}
    \SetKwInOut{KwIn}{Input}
    \SetKwInOut{KwOut}{Output}

    \KwIn{A TSP instance $V$, amount of edges to keep $k'$}
    \KwOut{A set of candidate edges}

    $min\_oneTree = \computeOneTree(V); \alpha = \dict(); candidateSet = \set()$

     \ForEach(\tcp*[h]{Iterate over all possible TSP edges}){pair $u,v  \in V, u \neq v$}{%
        \eIf{$(u,v) \in min\_oneTree$}{
            $\alpha\left[(u,v)\right] = 0$
         }{
            \tcp{Compute a new (possibly non-minimal) 1-Tree required to contain edge $(u,v)$}
            $oneTree = \computeOneTreeForcingEdge(V, (u,v))$ \\
            $\alpha\left[(u,v)\right] = \cost(oneTree)-\cost(min\_oneTree)$
         }
    }
    \ForEach(\tcp*[h]{Iterate over all TSP nodes}){$u  \in V$}{
        \For(\tcp*[h]{Choose neighbors with lowest $\alpha$}){k' smallest $\alpha\left[(u,v)\right]$ with $v  \in V$}{$candidateSet.add((u,v))$}
    
    }

    \KwRet{$candidateSet$}
    \caption{A simple algorithm for computing the candidate set (based on \cite{Helsgaun2000AnEI})} \label{alg:candset}
\end{algorithm}
\DecMargin{1em}
Note, that there is a much more elegant and quicker way to compute the $\alpha$ values instead of computing new 1-Trees. 
We refer the reader to \cite{Helsgaun2000AnEI} for details.
 Motivated by the promising performance of LKH, 1-Trees seem to be an obvious choice to make TSP graphs sparse as they also meet all our requirements:
\begin{enumerate}
    \item Only few edges for each node are required (where \cite{Helsgaun2000AnEI} states that keeping only the $k' = 5$ most promising edges for each node is sufficient in their test cases).
    \item 1-Trees (as a variety of spanning trees) are naturally connected, leading to connected sparse graphs.
\end{enumerate}
For comparison, we show the sparse versions of the graph in \cref{fig:knn_sparse_small} when sparsified with the 1-Tree method instead of $k$-nn in \cref{fig:onetree_sparse_small}. 
Note that the graph in this figure is now connected.
Furthermore, we note that in our experiments (\cref{perf_spars} and \cref{experiments}) we will make the candidate set symmetric, which means that if an edge $(u,v)$ is included, we also include $(v,u)$, as the euclidean TSP we consider is a symmetric problem (the solution is optimal, no matter in which ``direction'' we travel).
According to \cite{Helsgaun2000AnEI}, MSTs contain between $70 \%$ and $80 \%$ of an optimal TSP solution, which means we can interpret this number as a pessimistic lower bound on the amount of optimal TSP edges in a TSP graph sparsified by a 1-Tree approach.

In the following, we refer to the sparsification method used in the candidate set generation of the LKH as ``1-Tree''.
``1-Tree'' in combination with a certain $k$ indicates that the $k$ most promising edges according to the 1-Tree candidate set generation of LKH are kept for each node in the graph.

\subsubsection{On the Relationship of Sparsification and Attention Masking}
So far, we have discussed the idea of sparsifying TSP graphs for GNN encoders.
Many papers use transformer architectures as their encoders, however, e.g. \cite{kool2018attention}, \cite{kwon2020pomo} and \cite{jin2023pointerformer}.
Nevertheless, graph neural networks in the form of GATs are closely related to transformers or can even be interpreted as such \cite{joshi2020transformers}.%
\cite{kool2018attention} also notes that their model can be interpreted as a GAT (despite not using any message passing operators in the way they are implemented in the PyTorch Geometric \cite{Fey/Lenssen/2019} framework).
Therefore, we note that our proposed sparsification can also be used for transformer models by creating attention masks for the nodes in a TSP instance.
The proposed attention masks correspond to the adjacency matrices of the sparsified TSP graphs.
Similar to GNNs, these attention masks prevent a node from attending to other nodes that it is not connected to in the sparse TSP graph representation.

\subsection{A Sparsification Based Framework for Learning to Route}
\begin{figure*}[!t]
    \centering
    \includegraphics[width=0.9\linewidth]{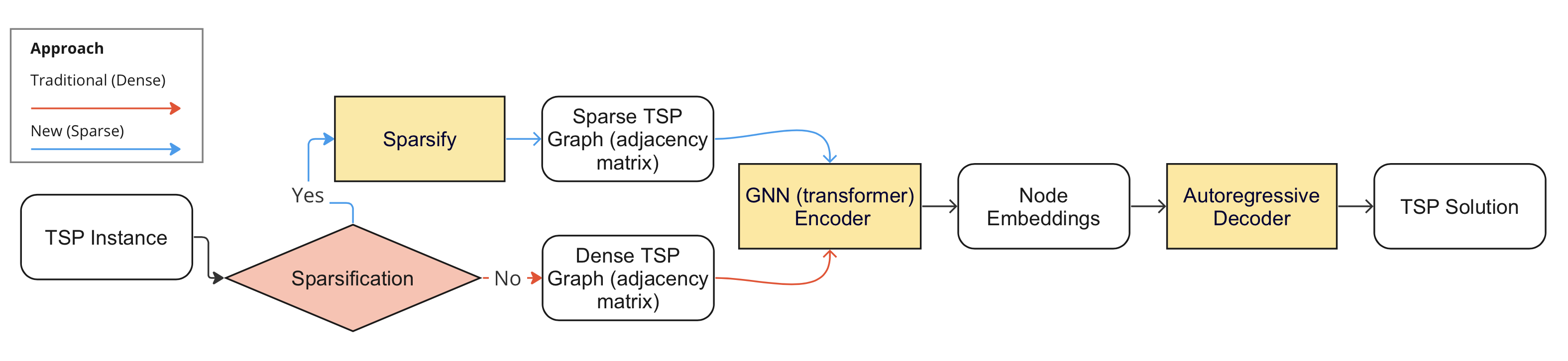}
    \caption{Flowchart presenting the overall encoder-decoder framework used in the experiments including new sparsification step (blue/top).}
    \label{fig:flowchart}
\end{figure*}

As pointed out in \cref{related_work}, virtually all learning based frameworks use an encoder architecture to capture the instance of the routing problem.
Even though all of these GNN and transformer based encoder architectures could profit from our sparsification method, we focus on encoder-decoder based approaches in our study to incorporate sparsification in the overall learning framework (our code is based on the framework of \cite{jin2023pointerformer}).
We visualize the overall framework in \cref{fig:flowchart}.

Previous studies would typically choose the ``red path'' shown in \cref{fig:flowchart} which means passing the dense TSP graph to the encoder.
We, on the other hand, propose the intermediate sparsification step (``blue path'') before passing the TSP instances to the encoder.
We note that our framework is compatible with GNN and transformer encoders.
The following decoder architecture in our framework then uses the embeddings produced by the encoders as in traditional approaches.
 
\subsubsection{Ensembles of Different Sparsification Levels}
\begin{figure*}[!t]
\centering
\subfloat[Spasification by $3$-nn]{
  \includegraphics[width=.2\linewidth]{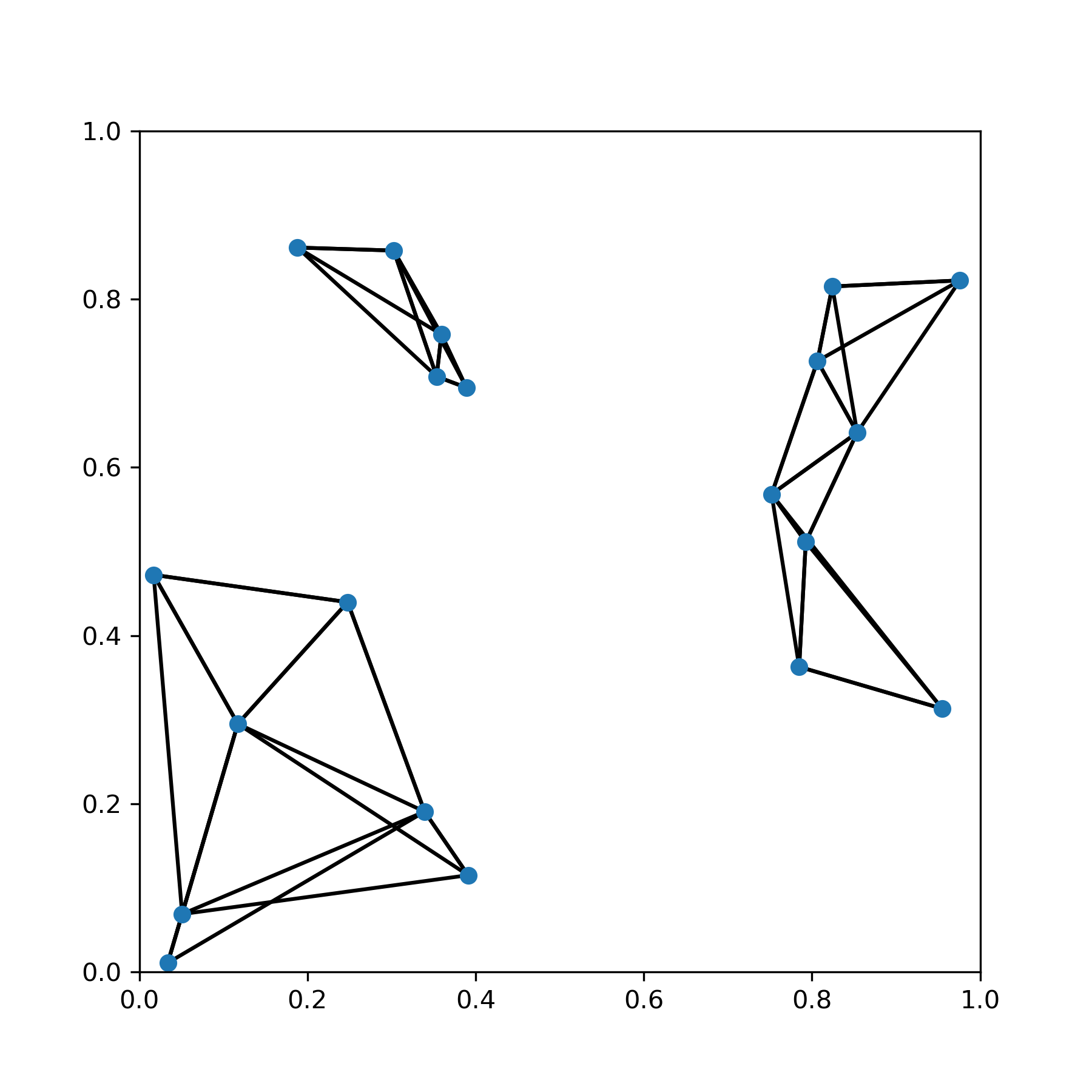}
  \label{fig:ensemble_3}
}
\hfil
\subfloat[Sparsification by $5$-nn]{
  \includegraphics[width=.2\linewidth]{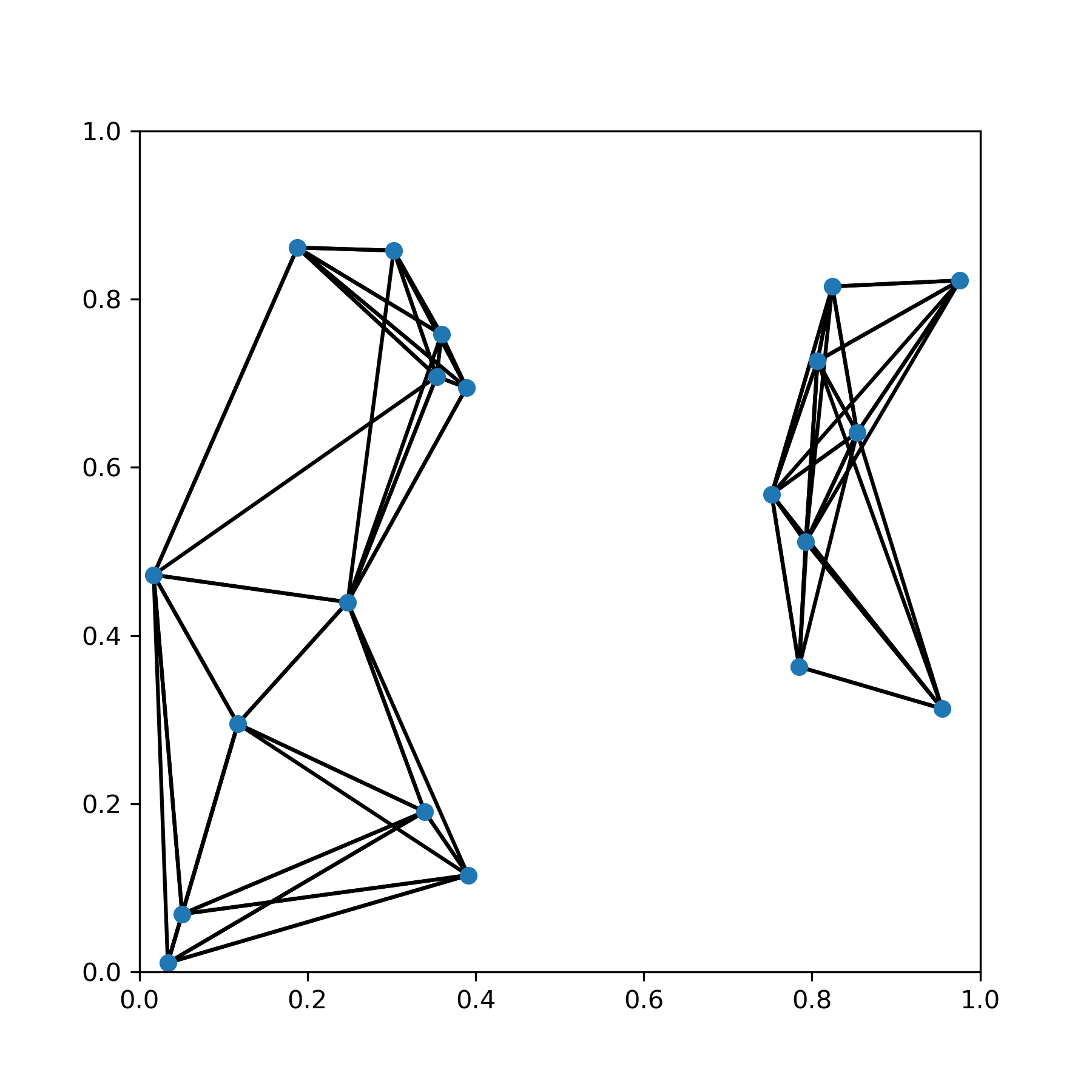}
  \label{fig:ensemble_5}
}
\hfil
\subfloat[Sparsification by $10$-nn]{
  \includegraphics[width=.2\linewidth]{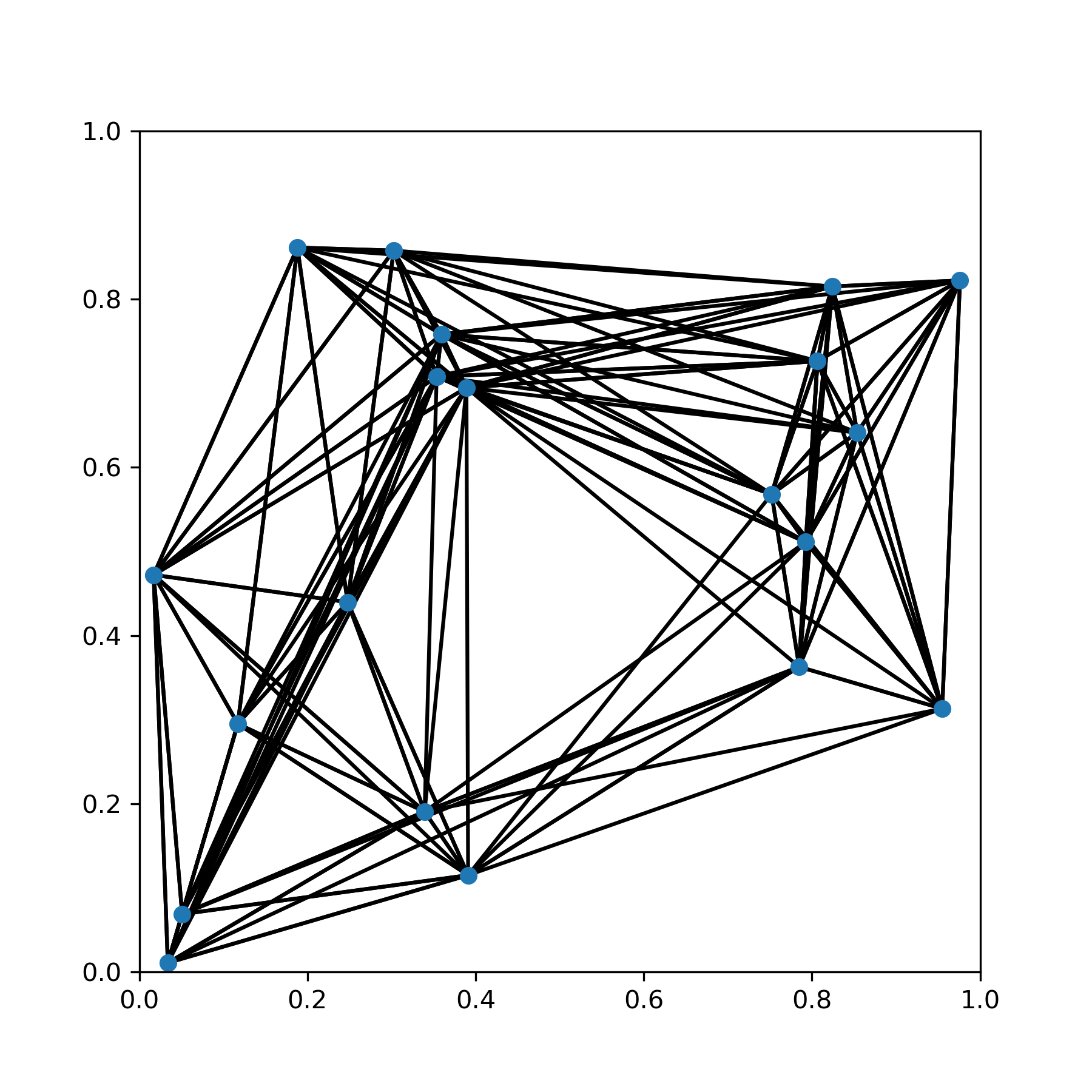}
  \label{fig:ensemble_10}
}
\hfil
\subfloat[Optimal TSP edges]{
  \includegraphics[width=.2\linewidth]{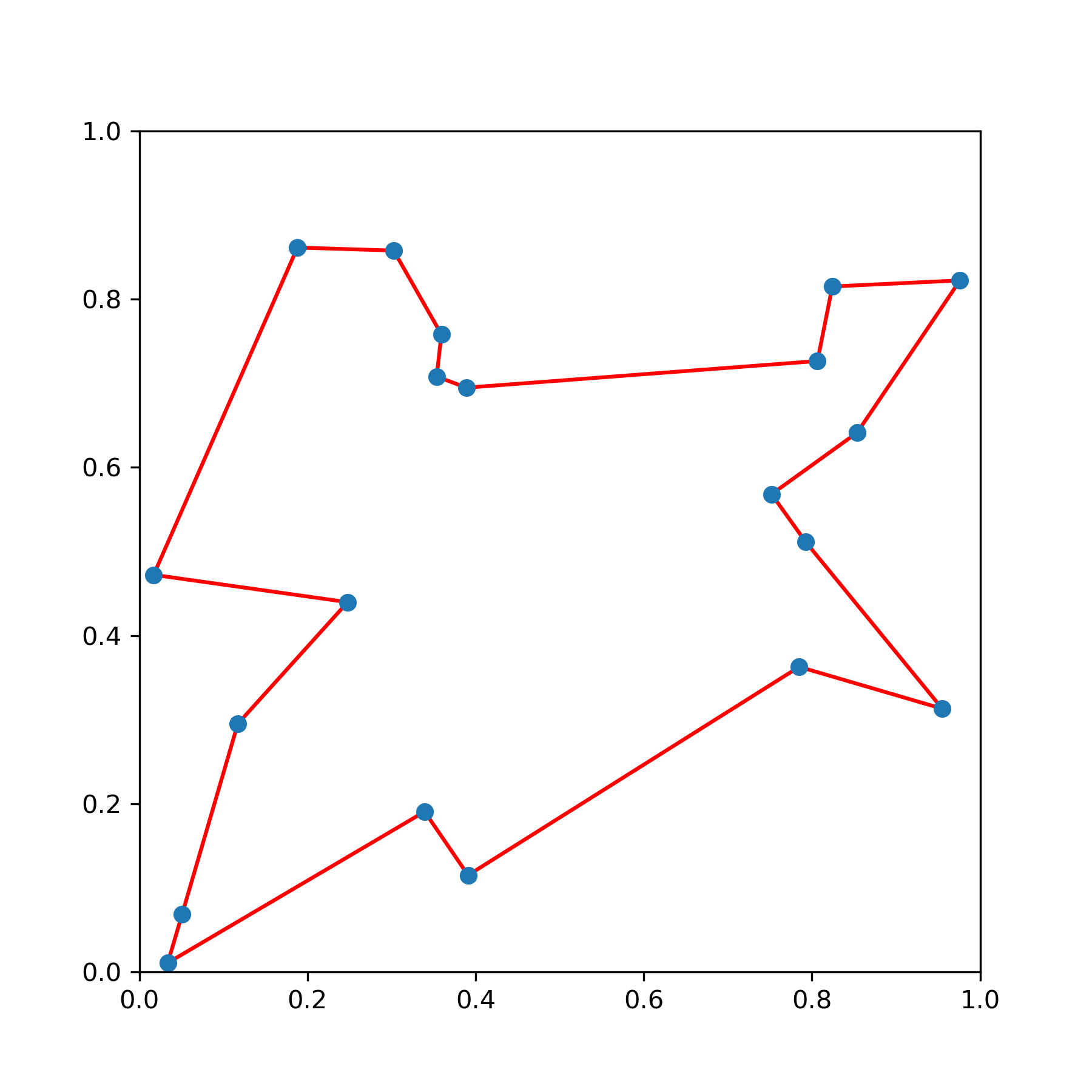}
  \label{fig:ensemble_solution_edges}
}
\caption{Ensemble Idea}
\label{fig:ensemble_figure}
\end{figure*}

As an additional contribution, we propose encoder-ensembles of different sparsification levels.
The idea of the ensemble encoder is to have several encoders operating on different sparsification level of the same TSP instance to provide a trade-off between information flooding while minimizing the risk of deleting optimal edges in sparse graph representations. 
We provide an example of this in \cref{fig:ensemble_figure}. 
In, \cref{fig:ensemble_3}, the graph is sparsified with $3$-nn.
This leads to a small number of overall edges, resulting in limited information flooding. 
However, we can see that the sparse graph is not connected and, therefore, not all edges of the optimal TSP solution are included in this sparse graph (compare \cref{fig:ensemble_solution_edges}).
Similarly, we provide a graph sparsified with $5$-nn in \cref{fig:ensemble_5} and a graph sparsified with $10$-nn in \cref{fig:ensemble_10}.
We note that \cref{fig:ensemble_10} provides a connected sparse graph that also includes all the optimal edges, whereas there were still optimal edges missing for $5$-nn in \cref{fig:ensemble_5}.
By combining the feature vector representations computed by several encoders in parallel on graphs of different sparsification levels, we hope to allow the overarching architecture to find the most important edges of the TSP instance, while minimizing the risk of completely deleting optimal edges. 

Overall, a TSP instance is processed in the ensemble framework as follows:
Several graph representations of different sparsification levels of the instance are generated.
Each graph representation is passed to a separate encoder, trained for the specific sparsification levels.
The encoders generate embeddings for all the nodes in the graph representations.
All embeddings corresponding to a specific node generated by the different encoders are concatenated and passed through a multilayer perceptron to create a single hidden feature vector representation of each node again.
The resulting node embeddings can be used in the decoder as usual.

\subsection{Other Sparsification Methods}
We note that in \cite{wang2018method} another method to sparsify TSP graphs was proposed which was based on frequency quadrilaterals.
The method works iteratively and removes $1/3$ of the edges in the graph in each iteration and, unlike our sparsification methods, it gives theoretical guarantees.
The authors report that for graphs with large $n$ (i.e., TSP instances with $n$ cities), two sparsification rounds can be performed without losing optimal edges.
This means that $\frac{2}{3} \cdot \frac{2}{3} \sim 44 \%$ of the edges remain in the sparse graph.  
Furthermore, the authors report that it is possible to perform $k$ sparsification rounds with $k \geq 2 + \lfloor log_{\frac{2}{3}} (\frac{1}{m}) \rfloor$ if at most $m$ optimal edges should be lost.
This implies that for $m = \frac{9}{4} < 3 $ the sparsification can be performed for $k=4$ iterations while losing at most 3 optimal edges.
For $k=4$, $(\frac{2}{3})^4 \sim 20\%$ of the edges would remain.

We do not consider this sparsification method in our evaluations, however, as no code is publicly available and we hypothesize that the iterative nature of the procedure leads to a too high runtime and computational complexity for our setting.

\section{Experiments} \label{total_experiments}
In this section, we present our experiments.
First, in \cref{perf_spars} we investigate the capability of keeping optimal TSP edges in sparse TSP graphs when sparsifying the instances with different sparsification methods.
Then, in \cref{experiments} we give an overview of the experimental setting when using sparse TSP instances for learning.
In \cref{results} we discuss the performance of GNNs operating on sparsified graphs and afterwards, in \cref{results_transformers}, we discuss the results of transformers that incorporate sparsification based attention masking.

\subsection{Optimal Edge Retention Capability of Sparsification Methods} \label{perf_spars}
We now investigate how well the two different sparsification methods perform in regards of keeping the optimal TSP edges in the sparse graph representation.
In particular, we select 100 random graphs of size 100 of two data distributions and keep the $k$ most promising edges for $k \in \{2, 3, \dots, 10\}$.
We then count for how many out of the 100 graphs all edges that are in the optimal solution of the TSP are also part of the sparse graph.
The two data distributions we investigate are \textit{uniform} and \textit{mixed}, explained in \cref{app:opt_edges}.
When evaluating the candidate edges generated with 1-Tree and $k$-nn on the 100 random graphs of the different data distributions, we obtain the results in \cref{Tab:Tcr}.
On the uniform data distribution, for $k=5$, 65 out of 100 sparse graphs contain all optimal edges when using 1-Tree for the sparsification.
This is considerably more than for $k$-nn, where only for 5 instances all optimal edges are in the sparse graph representation, but far away from the desired 100 instances.
For $k=10$, 98 sparse graphs contain all optimal edges if the sparsification was performed with 1-Trees.
On the mixed data distribution, the gap between $k$-nn and 1-Tree gets even bigger, which is most likely due to the coordinates of the nodes in the underlying graphs being more clustered.

Overall, even while not providing sparse graphs with 100\% optimal edge coverage for small $k$ like $k=5$, we expect 1-Tree to be nevertheless a better sparsification method than $k$-nn, especially on non-uniform, clustered data (which is more similar to real-world data) and for smaller, fixed $k$.\footnote{We note that we can adapt the hyperparameters (\hbox{``ASCENT\_CANDIDATES''} and \hbox{``INITIAL\_PERIOD''} in \cite{Helsgaun2000AnEI}) of the subgradient optimization in the 1-Tree generation slightly, to achieve sparse graphs with all optimal edges for all 100 graphs for $k=10$ at the cost of approximately double the runtime for computing the 1-Trees.
In particular, we can achieve this by doubling the ``ASCENT\_CANDIDATES'' hyperparameter and setting ``INITIAL\_PERIOD'' to 300.
We refer the reader to \cite{Helsgaun2000AnEI} for a detailed explanation of the influence of these hyperparameters.}
We believe that sparsifying the graphs and, therefore, deleting the majority of non-optimal edges will allow GNN encoders operating on TSP data to focus on the relevant parts of the problem, leading to better embeddings.

\begin{table}[!htb]
    \caption{Amount of graphs (out of 100) where the sparse graphs contain all optimal edges. The 65 in column ``1-Tree'', row ``5'' in the left table means that if the graphs are made sparse using 1-Trees and keeping the 5 most promising edges for each node in the graph, then for 65 graphs all edges of the optimal TSP solution are in the sparse graph, for 35 graphs there is at least one optimal edge missing.}
    \begin{minipage}{.5\linewidth}
      \centering
        \begin{tabular}{|c|c|c|}
         \hline
 \multicolumn{3}{|c|}{Uniform} \\
 \hline
k & NN & 1-Tree  \\ 
  \hline
2 & 0 & 0\\
  \hline
3 & 0  & 4 \\
  \hline
4 & 0  & 32  \\
  \hline
5 & 5  & 65 \\
  \hline
6 & 23  & 81\\
  \hline
7 & 48  & 91 \\
  \hline
8 & 62  & 93 \\
  \hline
9 & 73  & 97\\
  \hline
10 & 85  & 98 \\
 \hline
\end{tabular}
    \end{minipage}%
    \begin{minipage}{.5\linewidth}
      \centering
\begin{tabular}{|c|c|c|}

 \hline
 \multicolumn{3}{|c|}{Mixed} \\
 \hline
k & NN & 1-Tree  \\ 
  \hline
2 & 0 & 3\\
  \hline
3 & 0  & 14 \\
  \hline
4 & 0  & 48  \\
  \hline
5 & 2  & 76 \\
  \hline
6 & 12  & 90\\
  \hline
7 & 21  & 94 \\
  \hline
8 & 38  & 95 \\
  \hline
9 & 44  & 97\\
  \hline
10 & 49  & 98 \\
 \hline
\end{tabular}
\label{Tab:Tcr}
    \end{minipage} 
\end{table}

\subsection{Sparse Graphs for Encoders Evaluation} \label{experiments}
Here, we conduct two types of experiments, the first part is for GNN encoders to investigate the general idea of sparsificaiton and we aim to answer the following questions:
\begin{itemize}
    \item How does sparsification influence the performance of two different GNN encoders (GCNs and GATs) for different sparsification levels ($k$ values)?
    \item Is the best sparsification method dependent on the underlying datadistribution (uniform vs the more clustered \textit{mixed} distribution)?
    \item Can GATs learn the importance of nodes by themselves, making sparsification unnecessary?
    \item Can ensemble models of different sparsification levels lead to a tradeoff between focusing on the most important nodes only while also still allowing an information flow between all nodes in a graph (reducing the risk of unwantedly canceling out optimal connections in the TSP instance)?
    \item How do different dataset sizes influence the above questions?
\end{itemize}
To answer the above questions, we train combinations of GNN architectures (GAT, GCN), data distributions (uniform, mixed), sparsification methods ($k$-nn, 1-Tree), sparsification levels $k \in \{3,5,10,20,50\}$  and training dataset sizes $d \in \{500, 1000, 5000, 20000\}$ as encoders.
For the sake of brevity, we refer to such a combination as, e.g., (GAT, uniform, 1-Tree, $k$=3, $d$=500).
Moreover, we also train combinations like (GCN, mixed, dense, $d$=1000) to evaluate the performance of the GNNs on dense TSP graphs and combinations like (GCN, mixed, ensemble, $d$=1000) to test the ensembles of different sparsification levels.
The chosen degrees in the ensemble are $k=3,10,50$ to cover the range of different degrees as good as possible (compare \cref{ensemble_analysis}).
\newline

In the second experiment part, we aim to develop a new state-of-the-art model, which is based on transformer encoders (which have been used as encoders in many previous state-of-the-art frameworks).
In these experiments, we train a transformer ensemble of different sparsification levels ($k=3,20,50$) sparsified with the 1-Tree approach on the uniform data distribution (which has been used as the benchmark in the previous studies).
The sparsification is incorporated in the transformer encoders by sparsification-derived attention masking.

Transformers are trained on a dataset of size 1 million which was further increased via data augmentation (flipping and rotating instances) to create a dataset of size 16 million. This is necessary as the transformers suffer from severe overfitting on smaller datasets. In each epoch, 100,000 instances are chosen from the 16 million possible training instances (for 900 epochs total).
We note that our transformers stem from \cite{jin2023pointerformer} and were adjusted to support the sparsification based attention masking. In the original work of \cite{jin2023pointerformer} the transformers were trained on a dataset of 50 million instances (training was performed for 500 epochs and in each epoch 100,000 new instances were generated for training).
For the transformer, we also train ensemble and ``dense'' models on TSP instances of size 50 (with 16 million training instances again, but for 1000 epochs) to explore the generalizability of this model.
More information about the exact training settings can be found in \cref{setup_experiments}.

\subsection{Results - GNNs} \label{results}
We provide a table for each combination of a data distribution and GNN type, indicating the best optimality gap achieved in any epoch. %
All optimality gaps are w.r.t. the optimal solution generated by the concorde solver \cite{concorde}.
The tables can be interpreted as follows:
For a specific $k$ and $d$, the corresponding cell contains two values. 
The first one was achieved by using $k$-nn as the sparsification method, the second one was achieved by using 1-Tree.
Cells where 1-Tree achieves a better performance than $k$-nn are colored red, whereas cells where $k$-nn performs better than 1-Tree are colored blue.
White cells indicate that both initializations led to the same performance, or, in the case of ``dense'' cells, that no sparsification was performed.
Similarly, cells corresponding to ensembles ``ens'' are also colored to reflect whether the 1-Tree or $k$-NN ensemble worked better.
Furthermore, each row (corresponding to a certain dataset size $d$) has one bold and one underlined value.
The bold value corresponds to the very best optimality gap in the row, the underlined value to the second best score. 
\begin{table*}[!t]
\centering \setlength\tabcolsep{1.5pt}
\caption{Results - GAT - Optimality gap in percent}
\begin{tabular}{ |c|c|c|c|c|c|c|c| }
\hline
\multicolumn{8}{|c|}{Uniform - GAT} \\
\hline
\backslashbox{d}{k} & 3 & 5 & 10 & 20 & 50 & dense & ens \\
 \hline
500 & \cellcolor{red!10} 3.73/\underline{3.71} & \cellcolor{red!10} 3.82/\textbf{3.7} & \cellcolor{red!10} 4.8/4.42 & \cellcolor{red!10} 5.4/5.04 & \cellcolor{blue!10} 15.27/15.52 & 15.88 &  \cellcolor{red!10} 5.44/4.93  \\
  \hline
1000 & \cellcolor{red!10} \underline{2.77}/\textbf{2.66} & \cellcolor{red!10} 2.81/\underline{2.77} & \cellcolor{red!10} 2.92/2.88 & \cellcolor{red!10} 3.26/2.88 & \cellcolor{red!10} 5.47/5.2 & 14.5 & \cellcolor{red!10} 3.37/3.35 \\
\hline
5000 & \cellcolor{red!10} 1.7/1.51 & \cellcolor{blue!10} 1.55/1.56 & \cellcolor{red!10} 1.56/1.48 & \cellcolor{blue!10} \underline{1.43}/1.47 & \cellcolor{blue!10} 1.68/1.73 & 5.82 & \cellcolor{red!10} 1.61/ \textbf{1.42} \\
\hline
20000 & \cellcolor{red!10} 1.06/0.92 & \cellcolor{red!10} 0.89/0.84 & \cellcolor{red!10} 0.79/0.72 & \cellcolor{red!10} 0.72/\textbf{0.7} & \cellcolor{blue!10} 0.76/0.98 & 15.66& \cellcolor{red!10} 0.74/ \textbf{0.7} \\
\hline
  \multicolumn{8}{c}{} \\
  \hline
\multicolumn{8}{|c|}{Mixed - GAT} \\
\hline
  \backslashbox{d}{k} & 3 & 5 & 10 & 20 & 50 & dense & ens\\
 \hline
500 & \cellcolor{red!10} 5.63/\textbf{5.26} & \cellcolor{red!10} 5.74/\underline{5.53} & \cellcolor{blue!10} 5.93/6.21 & \cellcolor{red!10} 12.77/12.63 & \cellcolor{red!10} 14.39/14.29 & 14.53 & \cellcolor{red!10} 6.78/5.91 \\
\hline
1000 & \cellcolor{red!10} 4.55/\textbf{4.09} & \cellcolor{red!10} 4.6/\underline{4.39} & \cellcolor{blue!10} \underline{4.39}/4.52 & \cellcolor{red!10} 5.02/4.8 & \cellcolor{blue!10} 14.05/14.2 & 14.42 & \cellcolor{red!10} 5.34/4.49 \\
\hline 
5000 & \cellcolor{red!10} 2.61/2.13 & \cellcolor{red!10} 2.37/2.21 & \cellcolor{red!10} 2.24/2.15 & \cellcolor{red!10} 2.12/\underline{2.05} & \cellcolor{red!10} 2.07/ \textbf{2.02} & 14.32 & \cellcolor{red!10} 2.46/2.12 \\
\hline 
20 000 & \cellcolor{red!10} 1.7/1.31 & \cellcolor{red!10} 1.48/1.44 & \cellcolor{red!10} 1.39/1.32 & \cellcolor{blue!10} 1.31/1.36 & \cellcolor{red!10} 1.26/\underline{1.2} & 14.29 & \cellcolor{red!10} 1.34/ \textbf{1.15} \\
  \hline
\end{tabular}
\label{table_gat}
\end{table*} 

\textbf{Uniform - GAT}:
First, we note that the uniform distribution part of \cref{table_gat} is mostly reddish, indicating a tendency for 1-Tree to perform better. 
Furthermore, we note that the GNNs operating on dense graphs led to an overall performance that was much worse than any sparsified combination, no matter the dataset size.
Depending on the dataset size, the gap for the best sparse instances is between $\times 3$ to $\times 22$ better than the gap achieved on the dense graph representations. %
We hypothesize that this is due to two reasons:
First, the overall message passing architecture is not optimized for dense graphs.
Secondly, there is possibly not enough training data or time for the GNN to figure out via the attention mechanism which of the neighbors are important and which are not - even in the case of $20 000$ training instances.
We note that the best performance was achieved by the 1-Tree ensemble for $d=5000, 20000$, followed by moderately sparsified graphs with $k=20$.
On the smallest datasets, $k=3,5$ lead to the best results.

This aligns with our expectations that it is more difficult for the GAT to figure out the importance of nodes in settings with more edges (i.e. higher $k$) if the dataset is rather small. 
\newline
\textbf{Mixed - GAT}:
Like on the uniform dataset, the lower part of \cref{table_gat} is mostly redish, indicating a superiority of 1-Tree over $k$-nn .
For $d=20000$ the ensemble worked best again.
However, we note that $k=20$ and even $k=50$ work well on datasets of size $5000$ and $20000$.
On the smaller datasets, 1-Tree with $k=3,5$ leads to the best results.
We note that overall the performance of models operating on sparsified data is much better than on dense data.
We further point out that on the smaller datasets $k=50$ leads to a performance similarly bad as on dense data whereas $k=50$ leads to very good performance for $d=5000, 20000$. We hypothesize that this shows the capability of the GAT to learn important connections by itself via the attention mechanism if enough data is available.
\newline \textbf{Uniform - GCN}:
The upper part \cref{table_gcn} is completely red.
For all dataset sizes, the best results are achieved by the 1-Tree ensemble, followed by 1-Tree with $k=3$.
We note that for the GCN, the dense graphs lead to a rather reasonable overall performance, compared to the performance of dense GAT encoders.
Nevertheless, for $d=500$, the optimality gap improves from $4.06\%$ (dense TSP graphs) to $1.87\%$ (1-Tree ensemble).
This corresponds to a $\times 2$ performance increase.
For $d=20000$ the optimality gap improvement is even bigger, from $2.35 \%$ for dense TSP graphs to $0.77 \%$ for 1-Tree ensembles, a $\times 3$ improvement.
\newline
\textbf{Mixed - GCN}:
In this setting, the results (visible in the lower part of \cref{table_gcn}) are similar to the results produced by the GCN on the uniform distribution.
The table is overall red again, indicating that 1-Tree performed often better than $k$-nn. 
The best results are achieved by the ensembles (followed by 1-Tree with $k=3$) again for all dataset sizes.
The optimality gap improvements are also similar to the uniform data distribution case, with around $\times 2$ improvement for $d=500$ and a $\times 3$ improvement for $d=20000$ (comparing dense graph representations to the ensembles).
\begin{table*}[!t]
\centering \setlength\tabcolsep{1.5pt}
\caption{Results - GCN - Optimality gap in percent}
\begin{tabular}{ |c|c|c|c|c|c|c|c| }
\hline
\multicolumn{8}{|c|}{Uniform - GCN} \\
\hline
\backslashbox{d}{k} & 3 & 5 & 10 & 20 & 50 & dense & ens \\
 \hline
500 & \cellcolor{red!10} 2.94/ \underline{2.08} & \cellcolor{red!10} 3.1/3.07 & \cellcolor{red!10} 3.56/3.3 & \cellcolor{red!10} 3.84/3.6 & \cellcolor{red!10} 4.05/3.84 & 4.06 & \cellcolor{red!10} 2.63/ \textbf{1.87} \\
\hline
1000 & \cellcolor{red!10} 2.53/\underline{1.72} & \cellcolor{red!10} 2.75/2.67 & \cellcolor{red!10} 3.06/2.86 & \cellcolor{red!10} 3.29/3.13 & \cellcolor{red!10} 3.57/3.37 & 3.71 & \cellcolor{red!10} 2.27/\textbf{1.54} \\
\hline
5000 & \cellcolor{red!10} 1.93/\underline{1.21} & \cellcolor{red!10} 2.03/1.92 & \cellcolor{red!10} 2.3/2.17 & \cellcolor{red!10} 2.57/2.39 & \cellcolor{red!10} 2.84/2.58 & 2.84 & \cellcolor{red!10} 1.54/ \textbf{1.01} \\
\hline
20000 & \cellcolor{red!10} 1.41/ \underline{0.94} & \cellcolor{red!10} 1.64/1.47 & \cellcolor{red!10} 1.91/1.73 & \cellcolor{red!10} 2.12/1.95 & \cellcolor{red!10} 2.32/2.05 & 2.35 & \cellcolor{red!10} 1.16/\textbf{0.77} \\
\hline
  \multicolumn{8}{c}{} \\
  \hline
  \multicolumn{8}{|c|}{Mixed - GCN} \\
\hline
\backslashbox{d}{k} & 3 & 5 & 10 & 20 & 50 & dense & ens \\
 \hline
500 & \cellcolor{red!10} 4.46/\underline{2.97} & \cellcolor{red!10} 4.47/4.33 &\cellcolor{red!10} 4.71/4.65 & \cellcolor{red!10} 5.01/4.87 & \cellcolor{red!10} 5.47/5.15 & 5.53 & \cellcolor{red!10} 3.9/\textbf{2.61} \\
\hline 
1000 & \cellcolor{red!10} 3.71/\underline{2.27} & \cellcolor{red!10} 3.63/3.54 & \cellcolor{red!10} 3.93/3.74 & \cellcolor{red!10} 4.19/3.98 & \cellcolor{red!10} 4.47/4.3 & 4.58 & \cellcolor{red!10} 3.21/\textbf{1.93}  \\
\hline 
5000 & \cellcolor{red!10} 2.62/\underline{1.4} & \cellcolor{red!10} 2.62/2.32 & \cellcolor{red!10} 2.83/2.65 & \cellcolor{red!10} 3.06/2.84 & \cellcolor{red!10} 3.27/3.11 & 3.22 &\cellcolor{red!10}  2.19/\textbf{1.2}\\
\hline
20 000 & \cellcolor{red!10} 2.12/ \underline{1.04} &\cellcolor{red!10} 2.12/1.78 &\cellcolor{red!10} 2.29/2.12 &\cellcolor{red!10}  2.49/2.3 & \cellcolor{red!10}2.59/2.49 & 2.59 & \cellcolor{red!10} 1.72/\textbf{0.87} \\ 
  \hline
\end{tabular}
\label{table_gcn}
\end{table*}

\textbf{Summary}:
We summarize that for the GCN, the ensembles and smaller $k$ lead to the overall best results.
For the GAT there is a tendency for bigger $k$ and ensembles (but not dense graphs!) to lead to the best results for bigger datasets and for smaller $k$ on smaller datasets.
We note that dense GAT encoders perform unexpectedly badly.
We hypothesize that this is because all of our models are implemented in PyTorch Geometric \cite{Fey/Lenssen/2019} with actual message passing operations. 
Previous papers like \cite{kool2018attention} implemented their architectures in plain PyTorch \cite{paszke2017automatic} and they correspond to ``traditional'' transformers.
We further note that on the uniform dataset, GCNs outperform GATs for $d=500,1000$ and $5000$ and GATs only achieve a better performance than GCNs for $d=20000$.
On the mixed data distribution, GCNs perform better than GATs for all dataset sizes.
We also note that for extensive sparsification, i.e., for $k=3$, 1-Tree based data preprocessing always performs better than $k$-nn.
For bigger $k$ on the other hand, it is more arbitrary whether $k$-nn or 1-Tree based sparsification leads to better results for GATs (for GCNs, 1-Tree is always better).
This is expected, however, as the overlap between the two sparse graphs is most likely becoming bigger as $k$ grows, making the two sparsification methods more similar.
To conclude, we note that sparsification does increase the overall performance and ensemble methods can be a meaningful tradeoff between different sparsification levels, leading to the best performance for GCNs consistently and for GATs on bigger datasets.

\subsection{Results - Transformers} \label{results_transformers}
\begin{table*}[!ht]
\centering
\caption{Performance Transformer}
\begin{tabular}{|c|c||c|c||c|c|}
\hline
Approach & Paper & $n = 100$ & time & $n=50$ & time \\
\hline
\hline

Search-Based & \cite{fu2021} & 0.04 & 14min & 0.01 & 8m \\

\hline
Search-Based & \cite{joshi2019efficient} & 1.39 & 40min & 0.01 & 18m \\
\hline \hline
Improvement-Based & \cite{kim2021learning} & 0.54 & 12h & 0.02 & 7h\\
\hline
Improvement-Based & \cite{wu2021learning} & 1.42 & 2h & 0.20 & 1.5h \\
\hline \hline
Encoder-Decoder & \cite{kool2018attention} & 2.26 & 1h & 0.52 & 24m\\
\hline
Encoder-Decoder & \cite{kwon2020pomo} & 0.14 & 1min & 0.03 & 16s \\
\hline
Encoder-Decoder & \cite{jin2023pointerformer} (50 million) & 0.16 & 52s & 0.02 & 12s\\
\hline
Encoder-Decoder & Ours dense (16 million) & 0.25 & 1.52min & 0.05 & 27s\\
\hline
Encoder-Decoder & Ours 1-Tree, k=3 (16 million) & 0.19 & 1.52min & - & - \\
\hline
Encoder-Decoder & Ours 1-Tree, k=5 (16 million) & 0.29 & 1.5min & - & - \\
\hline
Encoder-Decoder & Ours 1-Tree, k=10 (16 million) & 0.25 & 1.52min & - & - \\
\hline
Encoder-Decoder & Ours 1-Tree, k=20 (16 million) & 0.17 & 1.52min & - & -\\
\hline
Encoder-Decoder & Ours 1-Tree, k=50 (16 million) & 0.34 & 1.7min & - & -\\
\hline
Encoder-Decoder & Ours 1-Tree, ens (16 million) & 0.10 & 1.78min & 0.00 & 34s \\
\hline
\end{tabular}
\label{transformer_performance}
\end{table*}

We show an overview of different state of the art papers with their corresponding performance and additionally the results of our experiments on transformers in \cref{transformer_performance}.
For comparison, we provide (aside from our ensemble) also results for individual sparsification levels and a dense model trained on our dataset.
We note, that the architecture of \cite{jin2023pointerformer} is the exact same one as in the row ``Ours dense''. 
The only difference is, that \cite{jiang2022learning} has a training set of size 50 million, while our training set was only 16 million instances big.
For instances of size $n=100$, this leads to a performance decrease from a gap of $0.16\%$ to $0.25\%$.
We note, however, that our transformers with 1-Tree based attention masking achieve a performance an optimality gap of $0.19\%$ and $0.17\%$.
This means that these two masked transformers lead to a performance almost as good as the performance of the original architecture of \cite{jin2023pointerformer}, despite using less than $1/3$ of the training data.
Moreover, we point out that our ensemble achieves a gap of $0.10\%$, an improvement of over $30$ percent despite using much less data for training. 
We hypothesize that our proposed ensemble would achieve even better performance if trained on an equally big dataset of 50 million instances.
We reiterate that our ensemble leads the performance table among the encoder-decoder based architectures.
We acknowledge that the search-based architecture of \cite{fu2021} achieves a better performance, however, only at the cost of a runtime that is approximately three times as big (note that we have to include the preprocessing time \cref{preproctimes} for the sparsification to our overall runtime). Additionally, \cite{min2023unsupervised} (a search-based approach) reports a gap of $0.00\%$ at the cost of double our runtime for $n=100$, however, as their code is not publicly available we cannot verify these results and, therefore, do not include them here.\newline
For $n=50$, all other architectures have a worse performance than our ensemble of different sparsification levels. %

\section{Conclusion and Future Work} \label{conclusion}

In this work, we propose two data preprocessing methods for GNN and transformer encoders when learning to solve the travelling salesman problem: $k$-nearest neighbors and a \hbox{1-Tree} based approach.
The aim of both methods is to delete unnecessary edges in the TSP graph making it sparse and allowing the encoder to focus on the relevant parts of the problem.
We analyse both sparsification methods from a theoretical point of view, pointing out that TSP instances sparsified with 1-Tree are always connected. 
Moreover, we validate on random instances that the 1-Tree based approach is less likely to delete optimal edges of the TSP tour when producing the sparse graph.
We show that graph neural networks can increase their performance when sparsification is performed in a data preprocessing step.
Similarly, transformer encoders can increase their performance when attention masking is performed were the attention  masks represent the adjacency matrices of the sparsified graphs.
To provide a trade-off between sparsification (which allows the model to focus on the most important connections only) and density (allowing an information flow between all components of a TSP graph) we propose ensembles of different sparsification levels.
Our proposed ensembles of different transformers achieve state-of-the-art performance on TSP instances of size 100 and 50, reducing the optimality gaps from $0.14\%$ to $0.10\%$ and $0.02\%$ to $0.00\%$, respectively, while having a better run time than improvement-based approaches.

We emphasize that our encoders are independent of the chosen learning paradigm.
Moreover, we note that our proposed data preprocessing is highly flexible and can be incorporated in many learning based frameworks to solve the TSP.
 We leave it open to feature work to design search-based or improvement-based approaches, as well as approaches dealing with scalability, using encoders taking advantage of our proposed data preprocessing.

In a future work, we plan on extending the proposed sparsification methods for routing problems that incorporate additional constraints like time windows.
Furthermore, we plan to investigate the possibilities of incorporating sparsification in VRPs that permit the repeated visitation of the same node, e.g., the depot in the CVRP.
Additionally, we aim to explore optimal ensemble compositions for sparsification more closely.

\section*{Acknowledgments}

This work was performed as a part of the research project ``LEAR: Robust LEArning methods for electric vehicle Route selection'' funded by the Swedish Electromobility Centre (SEC).
The computations were enabled by resources provided by the National Academic Infrastructure for Supercomputing in Sweden (NAISS) at Chalmers e-Commons partially funded by the Swedish Research Council through grant agreement no. 2022-06725.

{\appendices

\section{Data Distributions} \label{app:opt_edges}
As discussed in the experiments, we use two different data distributions to test our sparisification methods: \textit{uniform} and \textit{mixed}.
So far, the uniform data distribution has been used in most papers, tackling routing problems like the TSP or CVRP with machine learning.
In this data distribution, all coordinates $(x,y)$ of the nodes are sampled uniformly at random within the unit square, i.e. $x, y \in \left[0, 1\right]$. 

In the mixed data distribution (find examples in \cref{fig:mixed_dist}), we start with a uniform node distribution and afterwards apply 100 random mutation operators. 
The mutation operators that can be chosen from are \textit{explosion}, \textit{implosion}, \textit{cluster}, \textit{expansion}, \textit{compression}, \textit{linear projection} and \textit{grid}, presented in more detail in \cite{bossek2019evolving}.
The coordinates sampled from the mixed distribution are much more clustered than the coordinates sampled from the uniform distribution, which is more similar to real world data.

\section{Experiments - Setup} \label{setup_experiments}
Each combination of GNN type, data distribtuion, sparsification method, $k$ and dataset size $d$ is trained for 1000 epochs on an NVIDIA T4 GPU with 16GB of VRAM and we report the performance of the architecture on the validation set after each epoch.
Dense combinations of GAT on $d=20000$, as well as any experiments involving transformers are trained on a NVIDIA A40 GPU with 48 GB of VRAM.
The validation set is only used to report the current performance, it is completely independent of the training set and is also \textit{not} used to finetune any model parameters.

For a given data distribution, the same \textit{unprocesssed} validation set of $1000$ instances was used for the experiments with GNNs.
This means, that e.g. (GAT, uniform, 1-Tree, $k$=10, $d$=\hbox{20 000}) had the same unprocessed validation dataset as (GCN, uniform, $k$-nn, $k$=50, d=1000), the validation sets only differed after the preprocessing.
As a result, the reported optimality gaps for GNNs in the results section (\cref{results}) for different combinations are comparable, as long as the stem from the same data distribution, as the optimal solutions on the validation sets are the same.
For the transformers, the validation dataset was of size $10,000$ and we used the same publicly available dataset as \cite{jin2023pointerformer} and other authors to ensure performance comparability on our state-of-the-art model.

We note that the GAT consists of 6 layers and the GCN consists of 3 layers.
Moreover, the GAT uses additional edge feature vectors whose initialization are an encoding of the edge distances. 
We update the edge feature vectors in each layer by adding the current layers' hidden feature vectors of the edges and nodes, i.e. the feature vector of an edge $(i,j)$ in layer $\ell$ denoted as $e_{i,j}^\ell$ is updated as $e_{i,j}^{\ell+1} = e_{i,j}^\ell + x_i^{\ell+1} + x_j^{\ell+1}$, where $x_i^{\ell+1} + x_j^{\ell+1}$ are the feature vectors of node $i,j$ (the edge's endpoints) in layer $\ell +1$. 

The edge weights $e_{j,i}$ in the GCN (see \cref{gnn} for details) are computed in dependence of the initialization.
For 1-Tree based initializations, the edges $\alpha$ scores were used, for dense and $k$-nn based initializations the edge distances were used.
For a node $i$, the used $\alpha$ or distance edge scores $(j,i)$ were first normalized to the interval $\left[0, 1\right]$.
Afterwards, each normalized score $n_{j,i}$ was flipped by computing $1- n_{j,i}$. 
This was done in order to assign high scores to edges that previously has a low score and vice versa (a low distance or $\alpha$ score indicates a promising edge, which is why we want the corresponding $e_{j,i}$ score to be high).
Subsequently, we applied softmax to the resulting flipped scores, as we want low scores to remain low but not zero (as in this case, the score $e_{j,i} = 0$ would prevent information flow from node $j$ to $i$ completely). 
Moreover, by this, the edge scores $e_{j,i}$ for a node $i$ sum up to 1. We show an example of the overall procedure in Algorithm \ref{alg:edgeweights} for a node with $4$ neighbors in the (sparsified) graph. 

Our code is based on the publicly available code of \cite{jin2023pointerformer}.
We only substituted the transformer encoder with our GNNs or adapted transformers that incorporate attention masking and adapted the datasets to represent the sparse graphs with the corresponding preprocessing.
We did not change the decoder (i.e., nodes get autoregressively selected to form a valid solution) and learning is done by the original POMO inspired (see \cite{kwon2020pomo}) robust reinforcement learning setting.
As the decoder stays the same, it is possible to choose any node after any other node when decoding, no matter if the corresponding edge was part of the sparse graph or not.
This is necessary in order to always be able to find a Hamiltonian cycle in the graph which would not be possible on (e.g. but not restricted to) unconnected sparse graphs, compare \cref{fig:knn_sparse_small}. 
An overview of our framework including the sparsification process can be found in \cref{fig:flowchart}.

We note that we merely chose the architecture of \cite{jin2023pointerformer} because the used encoder-decoder architecture can be trained end-to-end and does not require any additional search algorithms for the solution generation at inference time.
Moreover, the used encoder-decoder approach allowed for easy substitution of the original encoder with our own encoders operating on sparse graphs without further adjustments to the overall framework.
We emphasize that we also could have chosen a search-based or an improvement-based approach to test our proposed data preprocessing, where the node embeddings generated by our encoders operating on the preprocessed sparse TSP instance could be used to generate edge probability heatmaps or to determine improvement operators. \newline
Our code will be publicly available once the paper is accepted.

\IncMargin{1em}
\begin{algorithm}[!ht]

    \SetKwFunction{zeroOneNormalize}{zeroOneNormalize}
    \SetKwFunction{newList}{newList}
    \SetKwFunction{softmax}{softmax}
    \SetKwInOut{KwIn}{Input}
    \SetKwInOut{KwOut}{Output}

    \KwIn{node $i$, with neighbor distances or $\alpha$ scores $scores = \left[0.1, 0.1, 0.2, 0.3 \right]$}
    \KwOut{edge weights $e_{j,i}$ for node $i$ in the GCN}

    \tcp{Normalize scores to 0-1 range}
    $normalized = \zeroOneNormalize(scores)$
    \tcp{$normalized = \left[0, 0, 0.5, 1 \right]$}

    $flipped = \newList()$ \\
    \tcp{Flip all scores so close neighbors have high scores}
    \ForEach( ){$n  \in normalized$}{
        $flipped.append(1-n)$ 
        \tcp{$flipped = \left[1, 1, 0.5, 0 \right]$}
    }
    \tcp{Ensure far away neighbors have non-zero score}
    $weights = \softmax(flipped)$
    \tcp{$weights = \left[0.3362, 0.3362, 0.2039, 0.1237\right]$}

    \KwRet{$weights$}
    \caption{Edge weight score computation - example} \label{alg:edgeweights}
    
\end{algorithm}
\DecMargin{1em}

\begin{figure}[!h] 
    \centering
    \includegraphics[width = 0.99\linewidth]{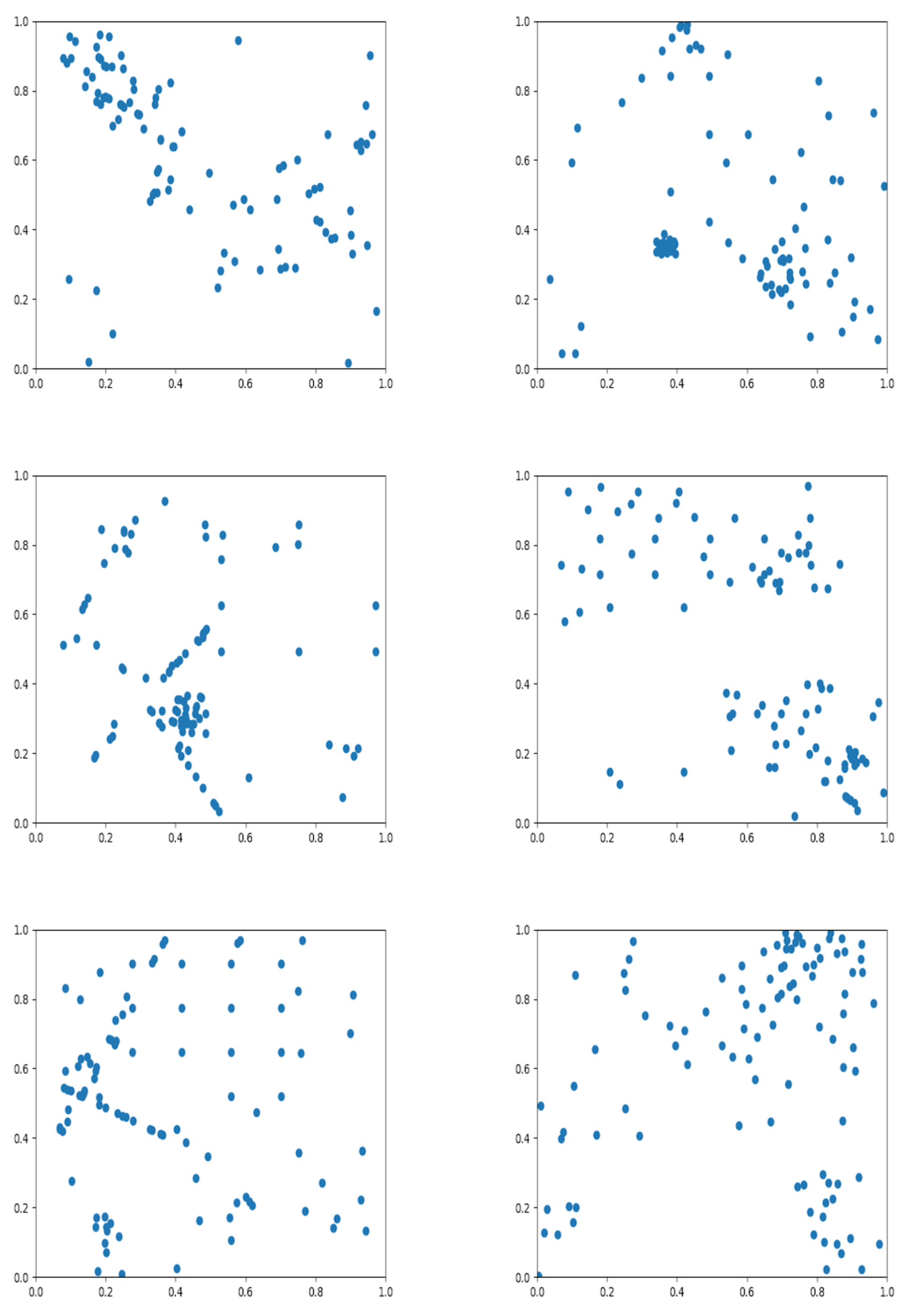}
  \caption{Visualization of 6 instances (of size 100) of the \textit{mixed} data distribution}
  \label{fig:mixed_dist} 
\end{figure}

\section{Data Augmentation}
The dataset for training the transformers was augmented from 1 million to 16 million (for TSP instances of size 50 and 100). %
This was done by flipping and rotating the instances.
We visualize this for an instance of size 4 in \cref{fig:flipping}.
We note that some coordinates close to the edges in the unit square might be outside of the unit square after rotating.
However, the model should be able to generalize to such cases and therefore we ignore this.
We performed data augmentation instead of additional instance generation as the preprocessing (sparsification) is time consuming for million of instances.
Therefore, we decided to preprocess an instance once and then use augment it by a factor of 16. We note that the optimal edges (and the edges after sparsification with $k$-NN and 1-Tree) do not change after a our data augmentation operations.

\begin{figure}[!ht]
    \centering
    \includegraphics[width = 0.5\linewidth]{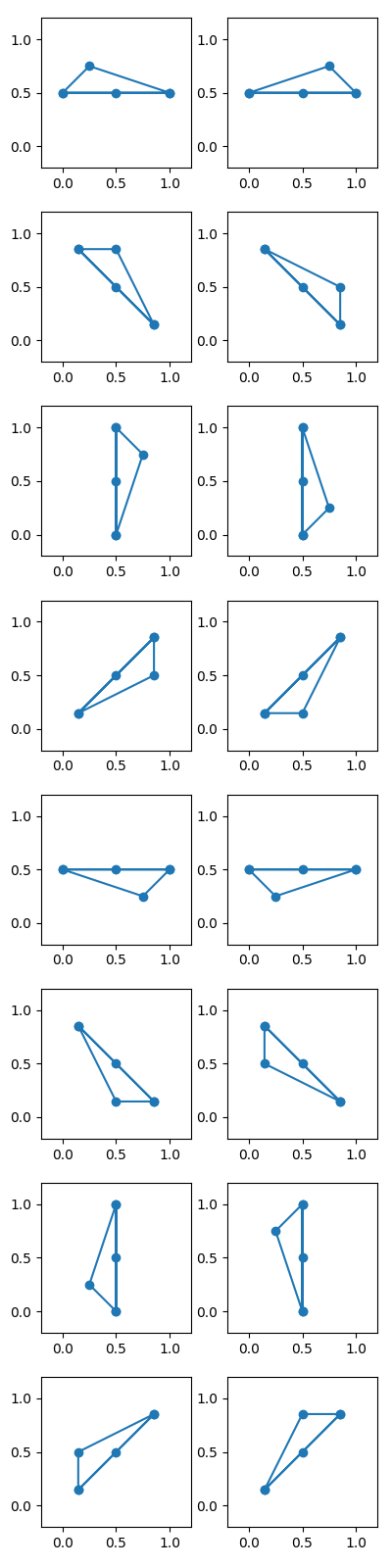}
    \caption{Data augmentation by flipping and rotating}
    \label{fig:flipping}
\end{figure}

\section{Experiments - Preprocessing Times}

We provide the times required for preprocessing 10000 instances with different (or no) sparsification methods.
We report the preprocessing times to compute ensembles of different sparsification levels ($k = 3,5,10,20, 50$) and including the additional computation time required for computing the edge weights used by the GCNs. 
We note that the preprocessing times for an individual, single $k$ can be much lower than for the complete ensemble of different $k$s.
All preprocessing times were achieved on a MacBook Air with M1 8-core CPU.
We note that preprocessing is done individually for each TSP instance and, therefore, it is highly parallelizable. 
As a result, using CPUs with more cores can decrease the runtime considerable.

\begin{table}[!htb]
\centering
\caption{Preprocessing times for 10 000 instances}
\begin{tabular}{ |c|c|c| }
\hline
initialization & nodes & time \\
  \hline
dense & 50 & 12s \\
\hline
$k$-nn & 50 & 40s \\
\hline
1-Tree & 50 & 2.2m \\
\hline
dense & 100 & 49s \\
\hline
$k$-nn & 100 & 1.7m \\
\hline
1-Tree & 100 & 3.6m \\
\hline 
\end{tabular}
\label{preproctimes}
\end{table}

\section{Experiments - Ensembles} \label{ensemble_analysis}

In order to investigate whether the models profited from ensembles of different sparsification levels, we also tested GNN ensembles of the same degree $k=3,3,3$ and $k=50,50,50$ compared to $k=3,10,50$ for datasets of size $20,000$.
These ensembles were also trained for 1000 epochs under the same settings as the mixed ensembles.
The results can be found in \cref{ensemble_study_table} and indicate that the models indeed leverage the different sparsification levels with the GCNs taking greater advantage than the GATs.

\begin{table}[!ht]
\centering \setlength\tabcolsep{1.5pt}
\caption{Results - Ensembles of the Same Sparsification Level}
\begin{tabular}{ |c|c|c|c| }
\hline
GNN & $k=3,3,3$ & $k=50,50,50$ & $k=3,10,50$  \\
\hline
GAT & 0.97 / 0.76 & 0.71 / 0.77 & 0.74 / \textbf{0.7} \\
\hline
GCN & 1.35 / 0.84 & 2.19 / 1.96 & 1.16 / \textbf{0.77} \\
\hline
\end{tabular}
\label{ensemble_study_table}
\end{table} 

}

\bibliographystyle{IEEEtran}

\newpage

\section{Biography Section}

\begin{IEEEbiography}[{\includegraphics[width=1in,height=1.25in,clip,keepaspectratio]{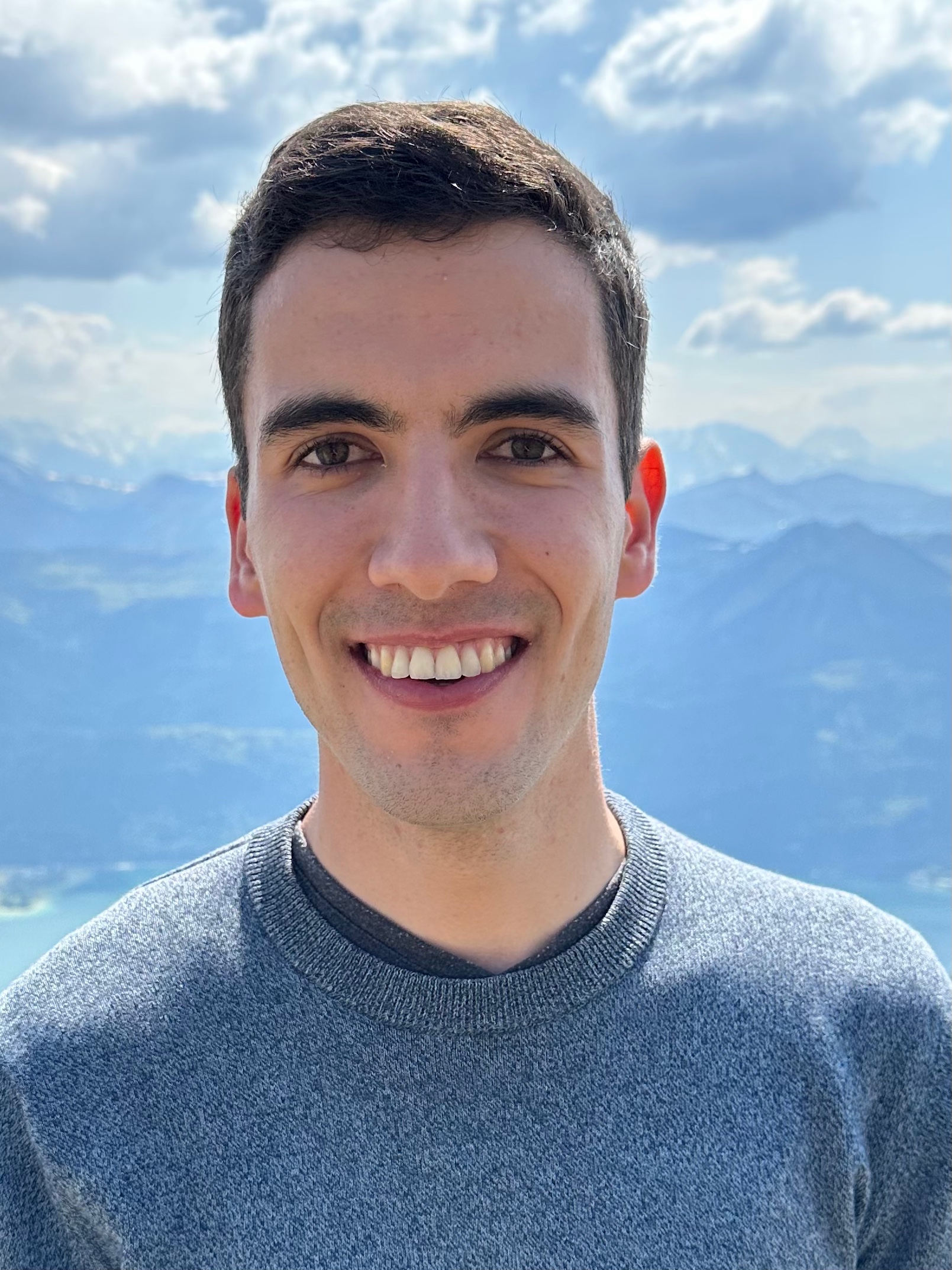}}]{Attila Lischka} received his B.Sc. and M.Sc. degree in computer science in 2021 and 2023 from RWTH Aachen University in Germany. His main research interests include machine learning, especially graph and reinforcement learning, as well as operations research and combinatorial optimization related topics such as routing problems.
\end{IEEEbiography}

\begin{IEEEbiography}[{\includegraphics[width=1in,height=1.25in,clip,keepaspectratio]{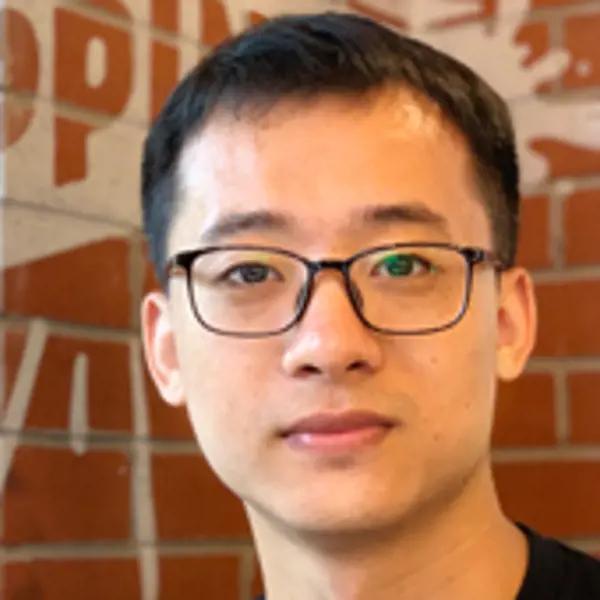}}]{Jiaming Wu} received his B.S. and M.S. degrees in the School of Transportation Science and Engineering from Harbin Institute of Technology in 2014, and his Ph.D. degree from Southeast University in 2019. He is currently a researcher in the Department of Architecture and Civil Engineering at Chalmers University of Technology, Gothenburg, Sweden. His research interests include electric vehicle fleet management (routing and charging), connected and automated vehicle platooning, and intersection control.
\end{IEEEbiography}

\begin{IEEEbiography}[{\includegraphics[width=1in,height=1.25in,clip,keepaspectratio]{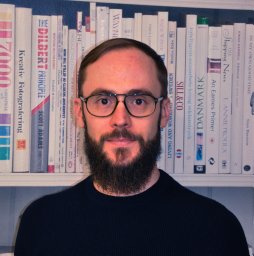}}]{Rafael Basso} received his PhD at Chalmers University of Technology in 2021 but employed by Volvo Trucks. He is currently working as a specialist with focus on energy consumption prediction and route planning for commercial vehicles using machine learning and physical models.
\end{IEEEbiography}

\begin{IEEEbiography}[{\includegraphics[width=1in,height=1.25in,clip,keepaspectratio]{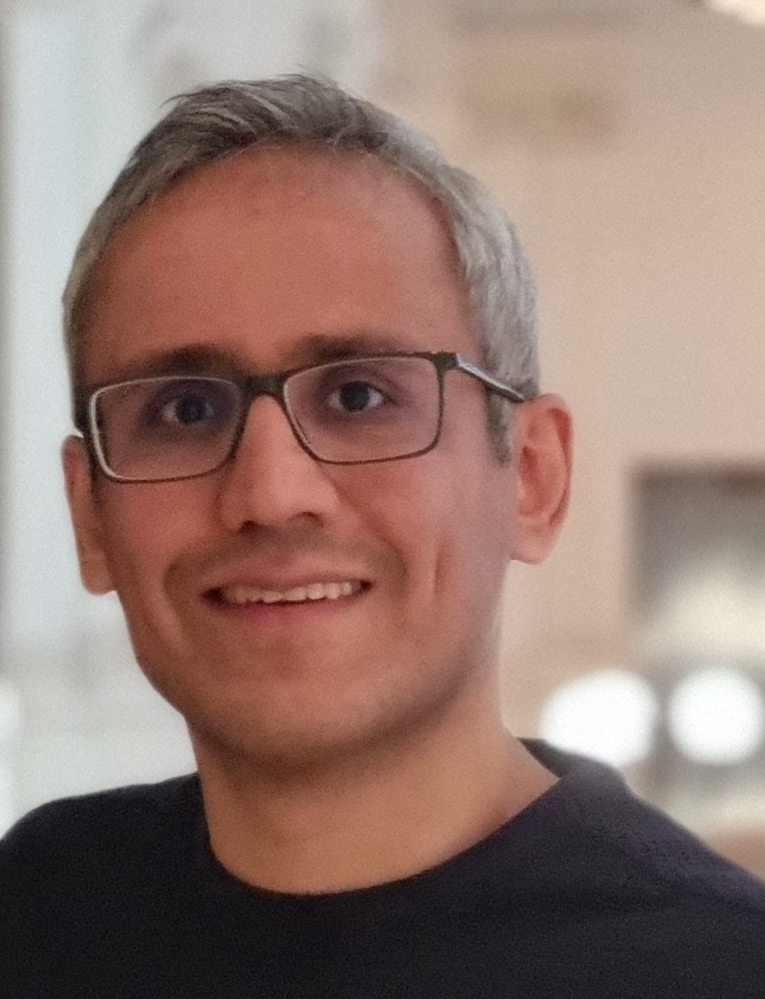}}]{Morteza Haghir Chehreghani} received his Ph.D. in computer science -- machine learning from ETH Zürich. After the Ph.D., he spent about four years with Xerox Research Centre Europe (later called Naver Labs Europe) as a Researcher (Staff Research Scientist I, and then Staff Research Scientist II). He joined Chalmers University of Technology in 2018, where he is an Associate Professor of machine learning, artificial intelligence, and data science with the Department of Computer Science and Engineering.
\end{IEEEbiography}

\begin{IEEEbiography}[{\includegraphics[width=1in,height=1.25in,clip,keepaspectratio]{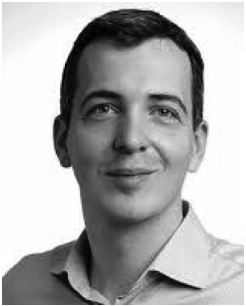}}]{Bal\'{a}zs Kulcs\'{a}r }  received the M.Sc. degree in traffic engineering and the Ph.D. degree from the Budapest University of Technology and Economics (BUTE), Budapest, Hungary, in 1999 and 2006, respectively. He has been a Researcher/Post-Doctor with the Department of Control for Transportation and Vehicle Systems, BUTE, the Department of Aerospace Engineering and Mechanics, University of Minnesota, Minneapolis, MN, USA, and the Delft Center for Systems and Control, Delft University of Technology, Delft, The Netherlands. He is currently a Professor with the Department of Electrical Engineering, Chalmers University of Technology, Göteborg, Sweden. His main research interest focuses on traffic flow modeling and control.
\end{IEEEbiography}

\vspace{11pt}

\vfill

\end{document}